\documentclass[10pt,twoside]{article}
\usepackage{amsmath}
\usepackage[utf8]{inputenc}
\usepackage{url}
\usepackage{dsfont}
\usepackage{graphicx}
\usepackage{yhmath}
\usepackage[table]{xcolor}
\usepackage{mathrsfs} 
\usepackage{amssymb}
\usepackage{url}
\usepackage{makecell}
\usepackage{arydshln}
\usepackage{multirow}
\usepackage{arydshln}
\usepackage{mathtools}
\usepackage{stmaryrd}  
\usepackage{booktabs, multirow}

\input epsf
\def\limiten{\renewcommand{\arraystretch}{0.5}
\begin{array}[t]{c}\stackrel{}{\longrightarrow} \\
{\scriptstyle n\rightarrow
\infty}\end{array}\renewcommand{\arraystretch}{1}}

\numberwithin{equation}{section}

\newtheorem{thm}{Theorem}[section]

\newtheorem{Def}[thm]{Definition\rm}

\newtheorem{prop}[thm]{Proposition}

\newtheorem{rmrk}[thm]{Remark}

\newcommand{\R}{\ensuremath{\mathbb{R}}}
\newcommand{\Z}{\ensuremath{\mathbb{Z}}}

\newcommand{\N}{\ensuremath{\mathbb{N}}}

\newcommand{\cov}{\ensuremath{\mathrm{Cov}}}

\newcommand{\lip}{\ensuremath{\mathrm{Lip}}}
\definecolor{grisclair}{gray}{0.9}
\font\dsrom=dsrom10 scaled 1200
\def \ind{\textrm{\dsrom{1}}}
\DeclareMathOperator*{\argmin}{argmin}

\newcommand{\bchi}{\ensuremath{\mbox{\large$\chi$}}}

\renewcommand{\arraystretch}{.8}

\begin{document}
\title{\bf Deep learning for  $\psi$-weakly dependent processes}
 \maketitle \vspace{-1.0cm}

\begin{center}
      William Kengne \footnote{Developed within the ANR BREAKRISK: ANR-17-CE26-0001-01 and the  CY Initiative of Excellence (grant "Investissements d'Avenir" ANR-16-IDEX-0008), Project "EcoDep" PSI-AAP2020-0000000013
   } 
   and 
     Modou Wade \footnote{Supported by the MME-DII center of excellence (ANR-11-LABEX-0023-01)} 
 \end{center}

  \begin{center}
  { \it 
 THEMA, CY Cergy Paris Université, 33 Boulevard du Port, 95011 Cergy-Pontoise Cedex, France\\
  E-mail:   william.kengne@cyu.fr  ; modou.wade@cyu.fr\\
  }
\end{center}

 \pagestyle{myheadings}
 \markboth{Deep learning for $\psi$-weakly dependent processes}{ Kengne and Wade}

~~\\

\textbf{Abstract}:
In this paper, we perform deep neural networks for learning $\psi$-weakly dependent processes.
Such weak-dependence property includes a class of weak dependence conditions such as mixing, association,$\cdots$ and the setting considered here covers many commonly used situations such as: regression estimation, time series prediction, time series classification,$\cdots$
 The consistency of the  empirical risk minimization algorithm in the class of deep neural networks predictors is established. 
 We achieve the generalization bound and obtain a learning rate, which is less than $\mathcal{O}(n^{-1/\alpha})$, for all $\alpha > 2 $. 
  Applications to binary time series classification and prediction in affine causal models with exogenous covariates  are carried out.
  Some simulation results are provided, as well as an application to the US recession data.
\medskip
 
 {\em Keywords:} Deep neural networks, $\psi$-weakly dependence, ERM principle, generalization bound, consistency.

\section{Introduction}
Deep learning has attracted a considerable attention in the literature and has been applied with a great success in several fields such as, image processing \cite{krizhevsky2017imagenet}, speech recognition \cite{hinton2012deep} and in general, for AI (artificial intelligence) roles in industry.
This interest is motivated by the very good accuracy of the deep neural networks (DNNs) algorithms in numerous applications, even if the theoretical properties of these algorithms in many settings are not yet well studied.
 One of the important properties of DNNs is that, they can be used to properly approximate several classes of functions, univariate or multivariate, see for instance \cite{schmidt2019deep}, \cite{ohn2019smooth} or \cite{schmidt2020nonparametric}.
In a past few years, several researchers have contributed to understand the theoretical advantages of DNNs.
%
%
For some results with independent and identically distributed (i.i.d.) observations, among others papers, \cite{dziugaite2017computing}, \cite{ohn2019smooth}, \cite{bauer2019deep}, \cite{valle2020generalization},  \cite{schmidt2020nonparametric}, \cite{kim2021fast}. 
But the i.i.d. assumption does not hold in many real life applications such as: market prediction, signal processing, meteorological observations,  forecasting of medical bookings,$ \ldots $ There are many contributions  to deep learning with dependent or non-i.i.d. observations, see for instance \cite{chen2019bbs}, \cite{kohler2020rate},  \cite{kurisu2022adaptive}, \cite{ma2022theoretical} and the references therein.

\medskip

We consider $D_{n}= \left\{Z_{1}=(X_{1},Y_{1}), \cdots, Z_{n}=(X_{n},Y_{n})\right\}$ (the training sample) from a stationary and ergodic process $\left\{Z_{t}=(X_{t}, Y_{t}), t\in\mathbb{Z}\right\}$, which takes values in $\mathcal{Z}=\mathcal{X}\times\mathcal{Y}$, where $\mathcal{X}$ is the input space and $\mathcal{Y} $ is the output space. 
We deal with a class of deep neural networks predictors $\mathcal{H_{\sigma}}(L, B, N, F, S)$ (see (\ref{H_equ2})) and a loss function $\ell:\mathbb{R}\times\mathcal{Y}\to [0, \infty)$.
%
For any hypothesis function $h\in \mathcal{H_{\sigma}}(L, B, N, F, S)$, define the risk,
\[R(h)= \mathbb{E}[\ell(h(X_{0}), Y_{0})], \]
%
%
and the empirical risk (with respect to $D_n$),
\begin{equation} \label{def_emp_risk}
\widehat{R}_{n}(h)=\frac{1}{n}\sum_{i=1}^{n}\ell(h(X_{i}), Y_{i}).
\end{equation} 
The goal is to build from the observations $D_n$, a deep neural networks predictor $ \widehat{h}_{n} \in \mathcal{H}_{\sigma}(L, N, B, F,S)$ with a low risk, such that, for any $t\in\mathbb{Z}$, $\widehat{h}_{n}(X_{t})$ is averaged "close" to $Y_t$. 
In the sequel, we set $\ell(h, z)= \ell(h(x), y) \; \text{for all} \; z= (x, y)\in \mathcal{Z} = \mathcal{X}\times\mathcal{Y}.$ 

\medskip

Let $h_{\mathcal{H}_{\sigma}(L, N, B, F,S)}$ be a target neural network (assumed to exist) and defined by,
\begin{equation} \label{def_target_DNN}
h_{\mathcal{H}_{\sigma}(L, N, B, F,S)} = \underset{h\in\mathcal{H}_{\sigma}(L, N, B, F,S)}{\argmin R(h)},
\end{equation} 
and $\widehat{h}_{n}$ be the neural network obtained from the empirical risk minimization (ERM) algorithm,
\begin{equation}\label{def_ERM_DNN}
    \widehat{h}_{n} = \underset{h\in\mathcal{H}_{\sigma}(L, N, B, F,S)}{\argmin \widehat{R}_{n}(h)}.
\end{equation}
We consider the ERM principle and aim to study the suitability of the estimation of $h_{\mathcal{H}_{\sigma}(L, N, B, F,S)}$ by $\widehat{h}_{n}$. That is, the generalization capability  of the ERM algorithm, which is accessed by studying how $R(\widehat{h}_{n})$ is close to $R(h_{\mathcal{H}_{\sigma}(L, N, B, F,S)})$.
The ERM algorithm is said to be consistent within the class of DNNs predictors ${\mathcal{H}_{\sigma}(L, N, B, F,S)}$, if $R(\widehat{h}_{n}) -R(h_{\mathcal{H}_{\sigma}(L, N, B, F,S)}) = o_{P}(1)$.  

\medskip

For a learning algorithm, one wants to calibrate a bound of the generalization error for any fixed $n$ (non asymptotic property) and investigate its consistency (asymptotic property).
As pointed out above, there are many works based on the study of the theoretical properties of DNNs for i.i.d. observations.
But, the theoretical studies for dependent observations are still scarce to date. 
 \cite{kurisu2022adaptive} have considered non-penalized and sparse-penalized DNNs estimators for non-parametric time series regression under some mixing conditions.
A consistency rate for prediction error of DNNs for $\alpha$-mixing observations has been obtained by \cite{ma2022theoretical}.
The works above are developed for time series regression within mixing-types conditions and do not consider a general setting that includes, for instance, pattern recognition.

\medskip

 This new contribution considers DNNs for learning a $\psi$-weakly dependent process  $\{Z_{t}=(X_{t},Y_{t}), t\in\mathbb{Z} \}$, with  values in $ \mathcal{Z}= \mathcal{X} \times \mathcal{Y} \subset \mathbb{R}^{d}\times\mathbb{R}$ (with $d \in \N$), based on the training sample $ D_{n}=\left\{Z_{1}=(X_{1}, Y_{1}), \cdots, Z_{n}=(X_{n}, Y_{n})\right\} $; and we address the following issues. 

 \medskip
 
 \begin{itemize}
 \item[(i)] \textbf{Consistency of the ERM algorithm over the class of DNNs}. We establish the consistency of the EMR algorithm with general loss function, for the DNNs predictors with a broad class of activation functions, for learning $\psi$-weakly dependent observations. Many classical models such as ARMAX, TARX, GARCH-X,$\ldots$ (see Section \ref{sect_AC}) or count time series (see for instance Proposition 1 in \cite{diop2022general}) fulfill such weak dependence structure. 
 Also, this dependence concept is more general than the mixing condition, since it is well known that many $\psi$-weak dependent processes do not satisfy a mixing conditions, see for instance \cite{dedecker2007weak}.
 In this sense, the results obtained here are more general than the existing ones in the literature.

 \item[(ii)] \textbf{Generalization bound and convergence rate of DNNs}. The generalization bound over the class of DNNs predictors is derived, as well as the learning rate. This rate is less than $\mathcal{O}(n^{-1/\alpha})$, for all $\alpha > 2 $; which is then close to the usual $\mathcal{O}(n^{-1/2})$ obtained in the general setting of the i.i.d. case.   
 \item[(iii)] \textbf{Application to time series classification}. Application to binary classification of a class of weakly dependent processes is carried out. For this purpose, we deal with a Lipschitz surrogate loss function and establish the consistency of the ERM algorithm over the class of DNNs predictors, which enjoy the generalization bound obtained. Real data application to the US recession data is also considered.
 \item[(iv)] \textbf{Application to affine causal models with exogenous covariates}. This class includes many classical autoregressive models such as ARMAX, TARX, GARCH-X, APARCH-X (see \cite{francq2019qml}).
 For prediction problem, it is shown that, the results of the consistency of the ERM principle within the DNNs and the generalization bound are applied to this class.  
 \end{itemize}
 
  \medskip
 
The rest of the article is organized as follows. Section 2 introduces the class of DNNs that will be considered. Some notations and assumptions are set in Section 3. Section 4 focuses on the consistency  of the ERM algorithm within the class of DNNs predictors, as well as the generalization bounds and their convergence rates.
Application to binary classification of a class of weakly dependent processes is carried out in Section 5, whereas Section 6 considers the application to affine causal models with exogenous covariates.
 Some numerical results are displayed in Section 7 and the proofs of the main results are provided in Section 8.

\section{Deep Neural Networks}
Fitting a DNN requires the choice of  an activation function $ \sigma: \mathbb{R}\to\mathbb{R}$ and the network architecture $(L, \textbf{p}), $ where $L \in \N$ is the number of hidden layers and $ \textbf{p}=(p_{0},   \cdots,  p_{L+1}) \in \N^{L+2}$ a width vector.
A DNN with $(L, \textbf{p})$ network architecture is any function  $ h $ defined by: 
\begin{equation}\label{h_equ1}
      h: \mathbb{R}^{p_{0}}\to\mathbb{R}^{p_{L+1}}, \;   y\mapsto h(y) = A_{L+1}\circ\sigma_{L}\circ A_{L}\circ\sigma_{L-1}\circ\cdots\circ\sigma_{1}\circ A_{1}(y),
\end{equation} 
 where $ A_{j}: \mathbb{R}^{p_{j -1}}\to\mathbb{R}^{p_{j}} $ is a linear affine map defined by $ A_{j}(y):= W_{j} y + \textbf{b}_{j}, $ for given  $p_{j - 1}\times p_{j}$  weight matrix   $ W_{j} $   and an shift vector $ \textbf{b}_{j}\in\mathbb{R}^{p_{j}}, $  and $ \sigma_{j}: \mathbb{R}^{p_{j}}\to\mathbb{R}^{p_{j}} $ is a nonlinear activation map  defined by $ \sigma_{j}(z) = (\sigma(z_{1}), \cdots, \sigma(z_{p_{j}}) )^{'}$. 
$p_0$, $p_{L+1}$ are, respectively, the input and the output dimension.
 In the sequel, we will assume that the activation function is $ C_{\sigma}$-Lipschitz for some $ C_{\sigma} > 0$; that is, $|\sigma(y_{2}) - \sigma(y_{1})|\leq C_{\sigma} |y_{2} - y_{1}|$ for all $y_{1}, y_{2} \in \mathbb{R}$.
For a DNN $h$ defined as in (\ref{h_equ1}), set
 \begin{equation}\label{theta_h}
 \theta(h):= \left(vec(W_{1})^{'}, \textbf{b}^{'}_{1}, \cdots,  vec(W_{L + 1})^{'} , \textbf{b}^{'}_{L+1}\right)^{'},
\end{equation} 
where $vec(W)$ transforms the matrix $W$ into the corresponding vector by concatenating the vectors of the column. 

\medskip

 Let $\mathcal{H}_{\sigma, p_{0}, p_{L+1}}$ be the class of DNNs predictors with the activation function $\sigma: \mathbb{R}\to\mathbb{R}$, that take $p_{0}$-dimensional input to produce $p_{L+1}$-dimensional output.
 Since the process $\{Z_{t}=(X_{t},Y_{t}), t\in\mathbb{Z} \}$ is with  values in $ \mathcal{Z}= \mathcal{X} \times \mathcal{Y} \subset \mathbb{R}^{d}\times\mathbb{R}$, we will deal with the class  $\mathcal{H}_{\sigma, p_{0}, p_{L+1}}$ with $ p_{0} = d $ and $ p_{L + 1}=1$.
  For a DNN $h$, denote by depth($h$) and width($h$) respectively the depth and the width of $h$, that is, if $h$ is a network with architecture $(L, \textbf{p})$, then, depth($h$)=$L$ and width($h$) =$\underset{1\leq j\leq L}{ \max} p_{j}$.
 For any positive constants $ L, N,   B , F$ and $S$, we set
\begin{equation}\label{h_sigma_L_N_B}
\mathcal{H}_{\sigma}(L, N, B):= \big\{h\in\mathcal{H}_{\sigma, d, 1}, ~ \text{depth}(h)\leq L, \text{width}(h)\leq N, \|\theta(h)\|_{\infty}\leq B \big\},
\end{equation}
 \begin{equation}\label{H_equ2}
 \mathcal{H}_{\sigma}(L, N, B, F):= \big\{ h\in H_{\sigma}(L, N, B), ~ \|h\|_{\infty}\leq F \big\}, 
 \end{equation}
 and, the class of sparsity constrained DNNs with sparsity level $S>0$ by
  \begin{equation}\label{H_LNBFS}
 \mathcal{H}_{\sigma}(L, N, B, F, S):= \big\{ h\in H_{\sigma}(L, N, B, F), ~ \|\theta(h)\|_{0} \leq S \big\}, 
 \end{equation}
 where $\|x\|_\infty = \underset{1\leq i \leq q}{\max} |x_i|$,  $\|x\|_0=\sum_{i=1}^q \ind_{x_i \neq 0} $  for all $x=(x_1,\ldots,x_q)' \in \R^q$, $q \in \N$,
 and
 $\| h \|_\infty  $ stands for the sup-norm of the function $h$ (see below).
 Throughout the sequel, $\mathcal{H}_{\sigma}(L, N, B, F, S)$ (with $L,N,B,F, S \geq 0$) denotes the class of DNNs defined in (\ref{H_equ2}). 

 \section{Notations and assumptions}
Let $E_1, E_2$ be two separable Banach spaces equipped with norms $\| \cdot\|_{E_1}$ and $\| \cdot\|_{E_2}$ respectively. 
Let us set some notations for the sequel.

\begin{itemize}
\item For all $ x \in \R$, $(x)_{+}=\max(x, 0)$. 
\item For any function $h: E_1 \rightarrow E_2$ and $U \subseteq E_1$,
\[ \| h\|_\infty = \sup_{x \in E_1} \| h(x) \|_{E_2}, ~ \| h\|_{\infty,U} = \sup_{x \in U} \| h(x) \|_{E_2} \text{ and }\] 
 \[\lip_\alpha (h) \coloneqq \underset{x_1, x_2 \in E_1, ~ x_1\neq x_2}{\sup} \dfrac{\|h(x_1) - h(x_2)\|_{E_2}}{\| x_1- x_2 \|^\alpha_{E_1}} 
 \text{ for any } \alpha \in [0,1] .\]
 \item For any $\mathcal{K}>0$ and $\alpha \in [0,1]$, $\Lambda_{\alpha,\mathcal{K}}(E_1,E_2)$ (simply $\Lambda_{\alpha,\mathcal{K}}(E_1)$ when $E_2 \subseteq \R$) denotes the set of functions  $h:E_1^u \rightarrow E_2$ for some $u \in \N$, such that  $\|h\|_\infty < \infty$ and  $\lip_\alpha(h) \leq \mathcal{K}$.
When $\alpha=1$, we set  $\lip_1 (h)=\lip(h)$ and $\Lambda_{1}(E_1) =\Lambda_{1,1}(E_1,\R)$. 
\item  $\mathcal{F}(E_1, E_2)$ denotes the set of measurable functions from $E_1$ to $E_2$.
\item For any $h \in \mathcal{F}(E_1, E_2)$ and $\epsilon >0$, $B(h,\epsilon)$ denotes the ball of radius $\epsilon$ of $\mathcal{F}(E_1, E_2)$ centered at $h$, that is,
$B(h,\epsilon) = \big\{ f \in \mathcal{F}(E_1, E_2), ~ \| f - h\|_\infty \leq \epsilon \big\}$.
\item For any $\mathcal{H} \subset \mathcal{F}(E_1, E_2)$, the $\epsilon$-covering number $\mathcal{N}(\mathcal{H},\epsilon)$ of $\mathcal{H}$ is the minimal number of balls of radius $\epsilon$ needed to cover  $\mathcal{H}$; that is,
\[ \mathcal{N}(\mathcal{H},\epsilon)= \inf\Big\{ m \geq 1 ~ : \exists h_1, \cdots, h_m \in \mathcal{H} \text{ such that } \mathcal{H} \subset \bigcup_{i=1}^m B(h_i,\epsilon)    \Big\} .\]
\end{itemize}

 \medskip

Let us define the weak dependence structure, see \cite{doukhan1999new} and \cite{dedecker2007weak}. Consider a separable Banach space $E$.
\begin{Def}\label{Def}
 $ An\; E- valued \;process \;(Z_{t})_{t\in\mathbb{Z}}\; is\; said\; to\; be\; (\Lambda_{1}(E), \psi,\epsilon)-weakly\; dependent\; $
 $ if\; there\; exists\; a\;\\ function\; $
 $\psi:[0, \infty)^{2}\times\mathbb{N}^{2}\to[0, \infty)\;and\; $
 $a \;sequence\;\epsilon=(\epsilon(r))_{r\in\mathbb{N}}\;decreasing\; to\; zero\;at\;infinity\;such\;that,\;for\;\\
any\;g_{1}, \;g_{2}\in\Lambda_{1}(E)\;with\; g_{1}:E^{u}\to\mathbb{R}, \;g_{2}:E^{v}\to\mathbb{R}\;(u, v\in\mathbb{N})\;and\;for\;any\; u-tuple\;(s_{1}, \cdots, s_{u})\;and\;any\;v-tuple\; (t_{1}, \cdots, t_{v})\;with\; s_{1}\leq\cdots\leq s_{u}\leq s_{u} + r\leq t_{1}\leq\cdots\leq t_{v}, \; the\; following\; inequality\; is \;fulfilled: $
\[|\cov(g_{1}(Z_{s_{1}}, \cdots, Z_{s_{u}}),  g_{2}(Z_{t_{1}}, \cdots, Z_{t_{v}}))|\leq\psi(\lip(g_{1}),\lip(g_{2}),u,v)\epsilon(r). \]
\end{Def}
For example, following choices of $\psi$ leads to some well-known weak dependence structures. 
 \begin{itemize}
 \item $\psi \left(\lip(g_1),\lip(g_2),u,v \right)= v \lip(g_2)$: the $\theta$-weak dependence, then denote $\epsilon(r) = \theta(r)$;
 \item $\psi \left(\lip(g_1),\lip(g_2),u,v \right)= u \lip(g_1) + v \lip(g_2)$: the $\eta$-weak dependence, then denote $\epsilon(r) = \eta(r)$;
 \item $\psi \left(\lip(g_1),\lip(g_2),u,v \right)= u v \lip(g_1) \cdot \lip(g_2)$: the $\kappa$-weak dependence, then denote $\epsilon(r) = \kappa(r)$;
  \item $\psi \left(\lip(g_1),\lip(g_2),u,v \right)= u \lip(g_1) + v \lip(g_2) + u v \lip(g_1) \cdot \lip(g_2)$: the $\lambda$-weak dependence, then denote $\epsilon(r) = \lambda(r)$.   
 \end{itemize}
 %
%
%
In the sequel,  for each of the four  choices of $\psi$  above, set  
respectively, 
\begin{equation}\label{Psi_eq}
\Psi(u, v)=2v, \; \Psi(u, v)=u + v, \; \Psi(u, v)=uv, \; \text{and} \;  \Psi(u, v)=(u + v + u v)/2.
\end{equation}

\medskip
 
We consider the process process $\{Z_{t}=(X_{t},Y_{t}), t\in\mathbb{Z} \}$ with  values in $ \mathcal{Z}= \mathcal{X} \times \mathcal{Y} \subset \mathbb{R}^{d}\times\mathbb{R}$, the loss function $\ell:\mathbb{R}\times\mathcal{Y}\to [0, \infty)$, the class of DNNs $\mathcal{H}_{\sigma}(L, N, B, F,S)$ with the activation function $\sigma: \R \rightarrow \R $, and set the following assumptions.
 \begin{itemize}
 \item[\textbf{(A1)}]: There exists  a constant $C_{\sigma} > 0$ such that the activation function $\sigma\in\Lambda_{1,C_{\sigma}}(\mathbb{R})$.
 \item[\textbf{(A2)}]: There exists $\mathcal{K_{\ell}}>0$ such that, the loss function $\ell \in \Lambda_{1, \mathcal{K_{\ell}}}(\mathbb{R}\times\mathcal{Y})$ and $M={\sup_{h\in\mathcal{H}_{\sigma}(L, N, B, F,S)}}{\sup_{z \in \mathcal{Z}}}|\ell(h, z)|<\infty$.
 \end{itemize}
Under \textbf{(A2)}, one can easily see that,
\begin{equation}\label{G_equ3}
G:= \underset{h_{1}, h_{2}\in\mathcal{H}_{\sigma}(L, N, B, F,S), h_{1}\not=h_{2}}{\sup}\underset{z\in\mathcal{Z}}{\sup}\frac{|\ell(h_{1}, z)- \ell(h_{2}, z)|}{||h_{1} - h_{2}||_{\infty}}<\infty.
\end{equation}
%
%
Let us set now the weak dependence assumption.
\begin{itemize}
\item[\textbf{(A3)}]: Let $\Psi:[0, \infty)^{2}\times\mathbb{N}^{2}\to [0, \infty)$ be one 
 of the choices in (\ref{Psi_eq}). The process $\left\{Z_{t}=(X_{t}, Y_{t}),{t\in\mathbb{Z}}\right\}$  is stationary ergodic and   $(\Lambda_{1}(\mathcal{Z}), \psi, \epsilon)$-weakly dependent such that, there exists $L_{1}, \, L_{2}, \, \mu\ge 0$
satisfying
\begin{equation}\label{A_equ4}
\sum_{j\ge 0}(j+1)^{k}\epsilon_{j}\leq L_{1}L_{2}^{k}(k!)^{\mu}  \; \text{for all} \; k\ge 0.
\end{equation}
\end{itemize}
The assumptions \textbf{(A1)}–\textbf{(A3)} hold for many classical models. 
Details are provided in Sections 5 and 6.

\section{Consistency of the EMR algorithm and generalization bounds}
\subsection{Consistency of the EMR algorithm}
The following proposition provides uniform (over the class $\mathcal{H}_{\sigma}(L, N, B, F,S)$) concentration inequalities between the risk and its empirical version.
This proposition is an application of Theorem 3.2 and 3.4 in \cite{diop2022statistical}.
\begin{prop}\label{prop} ~ 
\begin{enumerate}
\item[\rm 1.] Assume that the conditions \textbf{(A1)}-\textbf{(A3)} hold. 
 For all $\varepsilon>0, \,  n\in\mathbb{N},$  we have
\begin{equation}\label{P_equ4}
 P\Big\{\underset{h\in\mathcal{H}_{\sigma}(L, N, B, F,S)}{\sup}[R(h)-\widehat{R}_{n}(h)]>\varepsilon \Big\}\leq\mathcal{N}\Big(\mathcal{H}_{\sigma}(L, N, B, F,S),\frac{\varepsilon}{4G}\Big)\exp\Big(-\frac{n^{2}\varepsilon^{2}/8}{A_{n} + B_{n}^{1/(\mu + 2)}(n\varepsilon/2)^{(2\mu + 3)/(\mu + 2)}}\Big),
\end{equation}
   for any real numbers $A_{n}$ and $ B_{n}$ satisfying,
 $A_{n}\ge\mathbb{E}\Big[\Big(\sum_{i=1}^{n}\Big(\ell(h(X_{i}),Y_{i}) - \mathbb{E}[\ell(h(X_{0}),Y_{0})]\Big)\Big)^{2}\Big]$ and $B_{n}=2ML_{2}\max(\frac{2^{4+\mu}nM^{2}L_{1}}{A_{n}},1)$.
 \item[\rm 2.]  Assume that \textbf{(A1)}-\textbf{(A2)} hold and that $(Z_{t})_{t\in\mathbb{Z}}$ is $(\Lambda_{1}(\mathcal{X}\times\mathcal{Y}),  
 \psi, \epsilon)$-weakly dependent with $\epsilon_{j}=\mathcal{O}(j^{-2})$.  For any $\nu\in [0, 1]$ and for $n$ large enough, we have for all $\varepsilon>0,$
\begin{equation}\label{P_equ5}
    P\left\{\underset{h\in\mathcal{H}_{\sigma}(L, N, B, F,S)}{\sup}[R(h) - \widehat{R}_{n}(h)]>\varepsilon\right\}\leq C_{3}\mathcal{N}\left(\mathcal{H}_{\sigma}(L, N, B, F,S),  \frac{\varepsilon}{4G}\right)\exp\left(\log\log n - \frac{n^{2}\varepsilon^{2}/4}{A_{n}^{'} + B_{n}^{'}(n\varepsilon/2)^{\nu}}\right).
    \end{equation}
   with $ A_{n}^{'}\ge\mathbb{E}\Big[\Big(\sum_{i=1}^{n}\Big(\ell(h(X_{i}), 
 Y_{i}) - \mathbb{E}[\ell(h(X_{0}), Y_{0})]\Big)\Big)^{2}\Big],$ $B_{n}^{'}=\frac{n^{3/4}\log n}{A_{n}^{'}}$ and some constant $ C_{3}> 0$.
\end{enumerate}
\end{prop} 
 
 \medskip
 
  \medskip

 As in Remark 3.3 in \cite{diop2022statistical} and, for instance, the second attempt of Proposition \ref{prop} above, one can get, 
\[    P\left\{\underset{h\in\mathcal{H}_{\sigma}(L, N, B, F,S)}{\sup}[\widehat{R}_{n}(h) - R(h)]>\varepsilon\right\}\leq C_{3}\mathcal{N}\left(\mathcal{H}_{\sigma}(L, N, B, F,S),  \frac{\varepsilon}{4G}\right)\exp\left(\log\log n - \frac{n^{2}\varepsilon^{2}/4}{A_{n}^{'} + B_{n}^{'}(n\varepsilon/2)^{\nu}}\right),  \]
 and that,
\begin{equation}\label{prob_abs_inq}
    P\left\{\underset{h\in\mathcal{H}_{\sigma}(L, N, B, F,S)}{\sup}|R(h) - \widehat{R}_{n}(h) | >\varepsilon\right\}\leq 2 C_{3}\mathcal{N}\left(\mathcal{H}_{\sigma}(L, N, B, F,S),  \frac{\varepsilon}{4G}\right)\exp\left(\log\log n - \frac{n^{2}\varepsilon^{2}/4}{A_{n}^{'} + B_{n}^{'}(n\varepsilon/2)^{\nu}}\right).
    \end{equation}
Moreover, under the second attempt of Proposition \ref{prop}  with $\epsilon_{j}=\mathcal{O}(j^{-\gamma})$ for some $\gamma >3$  and from \cite{hwang2013study}, we have for all $h\in\mathcal{H}_{\sigma}(L, N, B, F,S)$,
 \begin{equation}\label{E_equ6}
 \mathbb{E}\Big[\Big(\sum_{i=1}^{n}\Big(\ell(h(X_{i}), Y_{i}) - \mathbb{E}[\ell(h(X_{0}), Y_{0})]\Big)\Big)^{2}\Big]\leq C n,
 \end{equation}
 where $C>0$ is a constant.
 Thus, by choosing $A'_{n}=n C$, one gets from (\ref{prob_abs_inq}) and Proposition 1 in \cite{ohn2019smooth} that,  $\underset{h\in\mathcal{H}_{\sigma}(L, N, B, F,S)}{\sup}|R(h) - \widehat{R}_{n}(h)|=\ o_{P}(1)$.
 This establishes the consistency of the EMR algorithm within the class of the DNNs $\mathcal{H}_{\sigma}(L, N, B, F,S)$ for weakly dependent processes, under a weaker condition than \textbf{(A3)}.
 \subsection{Generalization bounds}
 Let us derive generalization bounds of the EMR algorithm under the weak dependence conditions over the class of DNNs predictors $\mathcal{H}_{\sigma}(L, N, B, F,S)$.
%
%
 In the following Theorem \ref{thm1}, denote $ C_{1} = 4M^{2}\Psi(1, 1)L_{1} \; \text{and} \;  C_{2}= 2ML_{2}\max(\frac{2^{3 + \mu}}{\Psi(1, 1)}, 1)$,  where $\Psi$  is one of the functions in $(\ref{Psi_eq})$, under the assumption \textbf{(A3)}. 
\begin{thm}\label{thm1}
 Assume that \textbf{(A1)}-\textbf{(A3)} hold. Let $\eta\in(0, 1)$ and $\alpha > 2$. Assume that
\begin{equation*}
n\ge\max\Bigg( n_0(L, S,\alpha, M, \mu, C_4), \left(\frac{C_{4}}{2M^{2}}(C_{6} - \log(\eta) - 2L(S+1)\log(2M))_{+}\right)^{ \frac{ \alpha (\mu+2)}{\alpha-2}  } \Bigg) 
\end{equation*}
where
$C_{4}= 4C_{1} + 8C_{2}^{1/(\mu + 2)}M^{(2\mu + 3)/(\mu + 2)}$,  $C_{6}=2L(S+1)\log(4GC_{\sigma}L(N+1)(B\lor 1))$ and $n_0(L, S,\alpha, M, \mu, C_4) \in \N$ defined at (\ref{def_n_0}).
\begin{enumerate}
\item[\rm(i)] With probability at least $1 - \eta,$ we have 
\begin{equation}\label{equ_i}
R(\widehat{h}_{n}) - \widehat{R}_{n}(\widehat{h}_{n})\leq\varepsilon_{1}(n, \eta, \alpha),
\end{equation}
where
$\varepsilon_{1}(n, \eta, \alpha)<\frac{2M}{n^{\frac{1}{\alpha(\mu+2)}}}$ for all $\alpha > 2$.
\item[\rm (ii)] With probability at least $1 - 2\eta,$  we have 
\begin{equation}\label{equ_ii}
R(\widehat{h}_{n}) - R(h_{\mathcal{H}_{\sigma}(L, N, B, F,S)})\leq\varepsilon_{1}(n, \eta, \alpha)+\varepsilon_{1}'(n, \eta),
\end{equation}
where  
\[\varepsilon_{1}'(n, \eta)=\Big[\frac{\log(1/\eta)}{C_{n, 1}} \Big]^{\mu + 2} \text{ and } C_{n, 1}=\frac{n^{2}}{4C_{1}n + 8C_{2}^{1/(\mu +2)}(nM)^{(2\mu +3)/(\mu + 2)}}.\]
\end{enumerate}
\end{thm}
 The bound in (\ref{equ_i}) evaluates the estimation of $R(\widehat h_n)$ by $\widehat{R}_{n}(\widehat{h}_{n})$, whereas the bound in (\ref{equ_ii}) assesses how close this risk $R(\widehat h_n)$ to the risk of the target neural network in $\mathcal{H}_{\sigma}(L, N, B, F,S)$. 
The learning rates in the bounds (\ref{equ_i}) and $(\ref{equ_ii})$ are less than $\mathcal{O}(n^{-\frac{1}{\alpha(\mu + 2)}}) \; \text{for all} \; \alpha > 2$.
The following theorem provides more faster rate, under a weaker condition than \textbf{(A3)}.

   
\begin{thm}\label{thm2}
Assume that \textbf{(A1)} -\textbf{(A2)}  hold and that $(Z_{t})_{t\in\mathbb{Z}} $ is $(\Lambda_{1}(\mathcal{X}, \mathcal{Y}), \psi, \epsilon)-$ weakly dependent with $\epsilon_{j}=\mathcal{O}(j^{-\gamma})$ for some $\gamma>3.$  Let $\eta\in (0, 1)$, $\nu\in (0, 1).$ and $\alpha > 2$. Assume that 
 \begin{equation}\label{n_equ8}
 n > \left(\frac{C_{5}}{2M^{2}}(C_{6} - \log(\eta / C_{3}) - 2L(S+1)\log(2M))_{+}\right)^{\alpha/(\alpha-2)}
 \end{equation}
where $C_{3}$ is given in (\ref{P_equ5}),  $C_{5}=4(C+M^{\nu}/C)$ with the constant $C$ defined in (\ref{E_equ6}). 
\begin{enumerate}
 \item[\rm (i)] With probability at least $1-\eta,$  we have  for n large enough
\begin{equation}\label{R_equ9}
R(\widehat{h}_{n})-\widehat{R}_{n}(\widehat{h}_{n})\leq\varepsilon_{2}(n, \eta, \nu, \alpha),
\end{equation}
where 
$\varepsilon_{2}(n, \eta, \nu, \alpha) < \frac{2M}{n^{\frac{1}{\alpha}}},$  for all $\alpha > 2$. 
 \item[\rm (ii)] With probability at least $1-2\eta,$ we have  for n large enough 
\begin{equation}\label{R_10}
 R(\widehat{h}_{n})-R(h_{\mathcal{H}})\leq\varepsilon_{2}(n, \eta, \nu, \alpha) + \varepsilon'_{2}(n, \eta, \nu),
\end{equation}
where
\[ \varepsilon'_{2}(n, \eta, \nu)=\left[\frac{\log(C_{3}\log n/ \eta)}{C'_{n, 2}}\right]^{\frac{1}{2}} \text{ and } ~   
  C'_{n, 2}=\frac{n^{2}}{nC+\log n \; n^{\nu-1/4}(2M)^{\nu}/C}. \]
\end{enumerate}
\end{thm}

\medskip

\noindent The learning rates in the bounds $(\ref{R_equ9})$ and $(\ref{R_10})$ is less than $\mathcal{O}(n^{-1/\alpha}) \; \text{for all} \;  \alpha > 2$. 

\begin{rmrk}\label{rmk1}
 The generalization bounds derived in \cite{diop2022statistical} cannot be applied here, since the class $\mathcal{H}_{\sigma}(L, N, B, F,S)$ does not satisfy, in general, the condition \textbf{(A4)} of these authors; unless $\mathcal{X}$ is compact and we deal with the H{\"o}lder space $\mathcal{C}^s$ with $s=1$. In this case, the convergence rate obtained in \cite{diop2022statistical} is $\mathcal{O}(n^{-1/(2+2d)})$.
 Hence, the rate obtained above is more efficient and does not depend on the input dimension.
 \end{rmrk}

\section{Application to binary classification}
\subsection{Binary classification of weakly dependent processes} \label{sub_bin_class}
Let $(X_{1}, Y_{1}),...,(X_{n}, Y_{n})$ be a  trajectory of a stationary and ergodic process $Z_{t}=\left\{(X_{t}, Y_{t}),{t\in\mathbb{Z}}\right\}$, where $X_{t}\in\mathcal{X}\subset\mathbb{R}^{d}$ is the input vector and $Y_{t}\in\left\{-1, 1\right\}$ the class label. 
We focus on classifier $h:\mathcal{X}\to\mathbb{R}$, that predict $Y_t\in \mathcal{Y} = \left\{-1, 1\right\}$ based on $\text{sign}(h(X_{t}))$,  where $\text{sign}(x)=\ind_{\left\{x\ge0\right\}} - \ind_{\left\{x < 0\right\}}$ for all $x \in \R$.
The aim is to construct a predictor that minimizes the classification  risk, based to the 0-1 loss,
\begin{equation} \label{risk_01}
R_{01}(h):= P\big(\text{sign}(h(X_{0})\not=Y_{0}) \big)=E\big[\ind_{\left\{Y_{0}\cdot \text{sign}(h(X_{0})) \leq 0\right\}} \big].
\end{equation}
But, the optimization problem in (\ref{def_ERM_DNN}) with the 0-1 loss is computationally very hard, due to the non convexity of this loss function. 
An alternative approach is to deal with a margin-based, called surrogate loss  $\phi:\mathbb{R}\to [0,\infty)$ , and focus on the minimization of the surrogate risk, defined by, 
\begin{equation}\label{h__f}
R_{\phi}(h)=E[\phi(Y_{0} h(X_{0}))].
\end{equation}
The surrogate empirical risk is given by,
\[ \widehat{R}_{n, \phi}(h)=\frac{1}{n}\sum_{i=1}^{n}\phi(Y_{i}h(X_{i})) \]
and the DNN estimator based on this risk is defined  by,
\begin{equation}\label{h_equ}
\widehat{h}_{n, \phi}=\underset{h\in\mathcal{H}_{\sigma}(L, N, B, F,S)}{\argmin}\widehat{R}_{n, \phi}(h).
\end{equation}
Note that, if the surrogate loss $\phi$ is convex and Lipschitz, this is the case for the loss function $\ell(u,y)=\phi(u y)$.
A widely used example in this setting is the hinge loss, defined by, %
$\phi(z)=\max(1- z ,0 )$, which is convex and Lipschitz.
\subsection{Example of binary time series prediction}\label{sub_bin_pred}
Let $(Y_{1}, \mathcal{X}_{1}), \cdots, (Y_{n}, \mathcal{X}_{n})$ be a trajectory of a stationary and ergodic process $(Y_{t}, \mathcal{X}_{t})_{t\in\mathbb{Z}},$  where $(\mathcal{X}_{t})_{t\in\mathbb{Z}}$ is a process of covariates with values in $\R^{d_{x}}$, $d_{x}\in\N$. The goal is to predict $Y_{n+1}$ from the observations $(Y_{n}, \mathcal{X}_{n}), \cdots, (Y_{n-p+1}, \mathcal{X}_{n-p+1}), $ with $p\in\N$. 
We perform the learning theory with DNNs functions developed above with, $X_{t}=\left((Y_{t-1}, \mathcal{X}_{t-1}), \cdots, (Y_{t-p}, \mathcal{X}_{t-p})\right), \mathcal{Y}=\left\{-1, 1\right\} $, $\mathcal{X}\subset (\mathbb{R}\times\mathbb{R}^{d_x})^{p},$ and a  
margin-based loss function $\ell:\mathbb{R}\times\mathcal{Y}\to[0, \infty)$, defined from the hinge loss $\phi$.
 We focus on the class of DNNs predictors $\mathcal{H}_{\sigma}(L, N, B, F,S)$, with $L,N,B,F,S \geq 0$.
 
\medskip

Denote by $\mathcal{F}_{t-1}=\sigma\left\{Y_{t-1}, \cdots; \mathcal{X}_{t-1}, \cdots\right\}$ the $\sigma-$field generated by the whole past at time $t-1$ and assume that,
\begin{equation}\label{Mod_equ1}
 Y_{t}| \mathcal{F}_{t-1}\sim 2\mathcal{B}(p_{t})-1 \;\text{with} \; 2p_{t}-1=E[Y_{t}|\mathcal{F}_{t-1}]=f(Y_{t-1}, Y_{t-2}, \cdots; \mathcal{X}_{t-1}, \mathcal{X}_{t-2}, \cdots),
\end{equation}
where $f$ is a measurable non-negative function with values in $[-1; 1]$, $p_t=P(Y_t=1|\mathcal{F}_{t-1})$ and  $\mathcal{B}(p_{t})$ is the Bernoulli distribution with parameter $p_{t}$. 
The predictor obtained from the EMR algorithm with the surrogate loss function $\phi$ is given by,
\[ \widehat{h}_{n, \phi}=\underset{h\in\mathcal{H}_{\sigma}(L, N, B, F,S)}{\argmin}\widehat{R}_{n, \phi}(h) .\]
For any $t\in\mathbb{Z}$, the prediction of $Y_t$, based on $\widehat{h}_{n, \phi}$ is $\widehat{Y}_{t}=\text{sign}(\widehat{h}_{n, \phi}(X_{t}) )$ and thus, $\widehat{Y}_{n+1}=\text{sign}(\widehat{h}_{n, \phi}(X_{n+1}) )$.
Let us impose an autoregressive-type structure on the covariates:
\begin{equation}\label{aut_equ}
 \mathcal{X}_{t}=g(\mathcal{X}_{t-1}, \mathcal{X}_{t-2}, \cdots; \eta_{t}),   
\end{equation}
where $(\eta_{t})_{t\in\mathbb{Z}}$ is a sequence of i.i.d. random vectors with values in $\R^{d_\eta}$ ($d_\eta \in \N$) and $g(x; \eta)$ is a measurable function with values in $\mathbb{R}^{d_{x}},$ satisfying\\
\begin{equation}\label{ge_equ}
    \mathbb{E}[\|g(0; \eta_{0})\|^r]< \infty  \; \text{and} \; ||g(x; \eta_{0} - g(x'; \eta_{0}))||_{r}\leq \sum_{k=1}^{\infty}\alpha_{k}(g)||x_{k}-x'_{k}|| \;\text{for all} \; x,x'\in (\mathbb{R}^{d_{x}})^{\infty},
\end{equation}
for some $r\ge 1$, and a non-negative sequence $(\alpha_{k}(g))_{k\ge 1}$ satisfying $\sum_{k=1}^{\infty}\alpha_{k}(g) < 1$; where $||U||_{r}:=(\mathbb{E}||U||^{r})^{1/r}$ for any random vector $U$. 
We consider model (\ref{Mod_equ1}), (\ref{aut_equ}), (\ref{ge_equ}) and set the following assumptions. 
\begin{itemize}
\item For any $(y, x)\in\mathbb{R}^{\infty}\times(\mathbb{R}^{d_{x}})^{\infty}$,  $|f(y; x)|< 1 $ and there exists two sequences of non-negative real numbers $(\alpha_{k, Y}(f))_{k\ge 1}$ and  $(\alpha_{k, \mathcal{X}}(f))_{k\ge 1}$  satisfying $\sum_{k=1}^{\infty}\alpha_{k, Y}(f)< 1$ and  $\sum_{k=1}^{\infty}\alpha_{k, \mathcal{X}}(f)< \infty$; such that, for all $(y, x), (y', x')\in\mathbb{R}^{\infty}\times(\mathbb{R}^{d_{x}})^{\infty}$, 
\begin{equation}\label{lip_cond_f}
 |f(y; x)- f(y'; x')|\leq \sum_{k=1}^{\infty}\alpha_{k, Y}(f)|y_{k} - y'_{k}| + \sum_{k=1}^{\infty}\alpha_{k, \mathcal{X}}(f) \| x_{k} - x'_{k} \|, 
 \end{equation}
where $\|\cdot|\|$ denotes any vector norm in $\mathbb{R}^{d_x}$. 
\item  (\ref{ge_equ}) and (\ref{lip_cond_f}) hold with,
\begin{equation}\label{max_equ}
    \sum_{k=1}^{\infty}\max\left\{\alpha_{k}(g), \alpha_{k, Y}(f)\right\} <1.
 \end{equation}
\end{itemize}

Since the distribution $2\mathcal{B}(p)-1 $ belongs to the one-parameter exponential family, in the same way as in the proof Proposition 3.1 in \cite{lamine2020density}, we can show that there exists a $\tau$-weakly dependent stationary, ergodic solution $(Y_{t},  \mathcal{X}_{t})_{t\in\mathbb{Z}}$ of (\ref{Mod_equ1}) satisfying $\| (Y_{0},  \mathcal{X}_{0}) \|_r < \infty$.
This $\tau$-weak dependence structure implies the $\eta, \theta$-weak dependence. Indeed, for all $j\ge 0$, we have $\eta(j)\leq\theta(j)\leq\tau(j)$, see \cite{dedecker2007weak}.
 Therefore, there exists a solution $(Y_{t}, \mathcal{X}_{t})_{t\in\mathbb{Z}}$ of (\ref{Mod_equ1}) which is $\theta$-weakly dependent.
  Moreover, from \cite{doukhan2008weakly}, we get as $j\to \infty$,
\begin{equation}\label{teta_bound_equ}
    \theta(j)\leq\tau(j)=\mathcal{O}\left(\underset{1\leq \iota \leq j}{\inf}\left\{ \alpha^{j/ \iota} + \underset{k\ge \iota+1}{\sum}\alpha_{k}\right\}\right),
\end{equation}
where $\alpha_{k}=\max\left\{\alpha_{k}(g), \alpha_{k, Y}(f)\right\}$ and $\alpha=\sum_{k\ge 1}\alpha_{k}$.

\medskip

 We assume that the activation function $\sigma$ is Lipschitz continuous; for instance, ReLU (rectified linear unit): $\sigma(x) = \max(0,x)$, or sigmoid: $\sigma(x) = 1/(1+e^{-x})$. Let us check the other assumptions of Theorem \ref{thm1} and Theorem \ref{thm2} for the class  models (\ref{Mod_equ1}).

\begin{enumerate}
\item[(i)] If the surrogate  loss function is Lipschitz and bounded then, \textbf{(A2)} holds. For instance, the hinge loss $\phi(z)=\max(0, 1-z)$) fulfills these conditions. 
\item[(ii)]  Under the condition (\ref{max_equ}), the process $(Y_{t}, \mathcal{X}_{t})_{t\in\mathbb{Z}}$ is $\theta$-weakly dependent with the coefficients $\theta(j)$ bounded as in (\ref{teta_bound_equ}). One can easily get that, $(X_{t}, Y_{t})_{t\in\mathcal{Z}}$ is also $\theta$-weakly dependent with the coefficients $\theta(j)$.
Let us consider the following cases.\\
$\bullet$ The geometric case. Assume that,
$$\alpha_{k, Y}(f) + \alpha_{k}(g)=\mathcal{O}(a^{k}) \; \text{for some} \; a\in [0, 1).$$
From (\ref{teta_bound_equ}), one can easily get (see in \cite{doukhan2008weakly}) $\theta(j)\leq\tau(j)= \mathcal{O}\left(\exp(-\sqrt{\log(\alpha)\log(a)j})\right),$  where $\alpha$ is given in (\ref{teta_bound_equ}). Thus,  (\ref{A_equ4}) holds with $\mu=2$ (see also Proposition 8 in \cite{doukhan2007probability}). 
Therefore, the condition \textbf{(A3)} in Theorem \ref{thm1} holds. \ \

$\bullet$ The Riemanian case. Assume that,
$$ \alpha_{k, Y}(f) + \alpha_{k}(g)=\mathcal{O}(k^{-\gamma}) \; \text{for some } \; \gamma > 1. $$
From (\ref{teta_bound_equ}), one can easily get (see also \cite{doukhan2008weakly}) $\theta(j)\leq\tau(j)=\mathcal{O}\left(\left(\frac{\log j}{j}\right)^{\gamma-1}\right).$  Thus,  if $\gamma>3$, then the condition  $ \epsilon_{j}=\mathcal{O}(j^{-2})$ in Theorem \ref{thm2} is satisfied.
\end{enumerate}

\section{Application to affine causal models with exogenous covariates} \label{sect_AC}
Let $(\mathcal{X}_{t})_{t\in\mathbb{Z}}$ be a process of covariates with values in $\mathbb{R}^{d_{x}},$ $d_{x}\in\mathbb{N}$.  We consider the class of affine causal models with exogenous covariates (see \cite{diop2022inference}) defined by \\
\textbf{Class} $\mathcal{AC-}X(\mathcal{M}, f):$  A process $\left\{Y_{t}, t\in\mathbb{Z}\right\}$ belongs to  $\mathcal{AC-}X(\mathcal{M}, f) $ if it satisfies: 
\begin{equation}\label{cov_equ}
    Y_{t}= \mathcal{M}(Y_{t-1}, Y_{t-2}, \cdots; \mathcal{X}_{t-1}, \mathcal{X}_{t-2}, \cdots)\xi_{t} + f(Y_{t-1}, Y_{t-2}, \cdots; \mathcal{X}_{t-1}, \mathcal{X}_{t-2}, \cdots),
\end{equation}
where $\mathcal{M}, f$: $\mathbb{R}^{\mathbb{N}}\times( \mathbb{R}^{d_{x}})^{\mathbb{N}}\to\mathbb{R}$ are two measurable functions, 
 and $ (\xi_{t})_{t\in\mathbb{Z}} $ is a sequence of centered i.i.d. random variable satisfying $ \mathbb{E}(\xi_{0}^{r}) < \infty $ for some $ r\ge 2 $  and $\mathbb{E}(\xi_{0}^{2})=1$.
 The class $\mathcal{AC-}X(\mathcal{M}, f)$ includes the classical affine causal models without exogenous covariates studied in \cite{bardet2009asymptotic},  \cite{bardet2012multiple}, \cite{kengne2012testing}, \cite{bardet2020consistent},  \cite{kengne2021strongly}; which is obtained from (\ref{cov_equ}) when $\mathcal X_{t} \equiv C $ for some constant $ C $.
 Well known models such as ARMAX, TARX,  GARCH-X, ARMAX-GARCH  APARCH-X (see \cite{francq2019qml}) belongs to the class $\mathcal{AC-}X(\mathcal{M}, f)$. 
The inference in the class $\mathcal{AC-}X(\mathcal{M}, f)$ in a semiparametric setting based on a quasi likelihood estimator has been carried out in \cite{diop2022inference}.

 \medskip

 In order to study the stability properties of model (\ref{cov_equ}),   \cite{diop2022inference} set following Lipschitz-type conditions on the functions  $f, \mathcal{M}$ or $\mathcal{M}^{2}$.
 Throughout this section,  0 denotes the null vector of  any vector space.  For $\Psi = f$ or $\mathcal{M}$, consider the following assumptions.
 
 \medskip

\noindent $\textbf{Assumption A}(\Psi):  |\Psi(0; 0) | < \infty $ and there exists two sequences of non-negative real numbers $ (\alpha_{k, Y}(\Psi))_{k\ge 1} $ and $ (\alpha_{k, \mathcal{X}}(\Psi))_{k\ge 1} $  satisfying  $\sum_{k=1}^{\infty}\alpha_{k, Y}(\Psi) < \infty,$ $\sum_{k=1}^{\infty}\alpha_{k, \mathcal{X}}(\Psi) < \infty $; such that for any $ (y, x),(y', x')\in\mathbb{R}^{\infty}\times(\mathbb{R}^{d_{x})^{\infty}}, $  
 \[ |\Psi(y; x)- \Psi(y'; x')|\leq\sum_{k=1}^{\infty}\alpha_{k, Y}(\Psi)|y_{k} - y'_{k}| + \sum_{k=1}^{\infty}\alpha_{k, \mathcal{X}}(\Psi)\|x_{k} - x'_{k}\| .\] 
 %
 %
 Regarding the cases of ARCH-X type models, the next assumption is set on $H=\mathcal{M}^{2}$.
 
  \medskip
 
\noindent \textbf{Assumption A}($H$): Assume that $f=0$, $|\mathcal{M}(0; 0)|< \infty $ and there exists two sequences of non-negative real numbers  $ (\alpha_{k, Y}(H))_{k\ge 1} $ and $ (\alpha_{k, \mathcal{X}}(H))_{k\ge 1} $ satisfying $\sum_{k=1}^{\infty}\alpha_{k, Y}(H) < \infty,$ 
  $\sum_{k=1}^{\infty}\alpha_{k, \mathcal{X}}(H) < \infty; $  such that for any $ (y, x),(y', x')\in\mathbb{R}^{\infty}\times(\mathbb{R}^{d_{x})^{\infty}}, $ 
   \[ |H(y; x)- H(y'; x')|\leq\sum_{k=1}^{\infty}\alpha_{k, Y}(H)| y_{k}^{2} - {y'}_{k}^{2} | + \sum_{k=1}^{\infty}\alpha_{k, \mathcal{X}}(H)\|x_{k} - x'_{k}\|. \]
 The convention that if $\textbf{A}(\mathcal{M})$ holds,  then $\alpha_{k, Y}(H)=\alpha_{k, \mathcal{X}}(H)=0 $ for all $k\ge 1$ and if \textbf{A}($H$) holds,  then $\alpha_{k, Y}(\mathcal{M})=\alpha_{k, \mathcal{X}}(\mathcal{M})=0$ for all $k\ge 1$ is made for the sequel. 
 We consider the model (\ref{cov_equ}) with the conditions (\ref{aut_equ}), (\ref{ge_equ}) on the covariates where $(\eta_{t})_{t\in\mathbb{Z}}$ in (\ref{aut_equ}) is such that, $(\eta_{t}, \xi_t)_{t\in\mathbb{Z}}$ is a sequence of i.i.d. random vectors; and assume that, for some $r \geq1$,  
 \begin{equation}\label{Lipf_equ}
    \sum_{k=1}^{\infty}\max\left\{\alpha_{k}(g), \alpha_{k, Y}(f) + \|\xi_{0}\|_{r}\alpha_{k, Y}(\mathcal{M}) + \|\xi_{0} \|_{r}^{2}\alpha_{k, Y}(H)\right\} < 1 .
 \end{equation}
Under the assumptions $\textbf{A}(f)$, $\textbf{A}(\mathcal{M})$ or $\textbf{A}(\mathcal{H})$, and (\ref{Lipf_equ}), there exists a $ \tau-\text{weakly}$ dependent stationary, ergodic and non adaptive solution $(Y_{t}, \mathcal{X})_{t\in\mathbb{Z}}$ of (\ref{cov_equ}) satisfying $ ||(Y_{0}, \mathcal{X}_{0})||_{r}< \infty $ (see \cite{diop2022inference} ).
This solution is $\theta-\text{weakly dependent}$ 
with coefficients $\theta(j)$ bounded as in (\ref{teta_bound_equ}), with
\[ \alpha_{k}= \max\left\{\alpha_{k}(g), \alpha_{k, Y}(f) + \|\xi_{0}\|_{r}\alpha_{k, Y}(\mathcal{M}) + ||\xi_{0}||_{r}^{2}\alpha_{k, Y}(H)\right\} < 1  \; \text{ and } \;  \alpha=\sum_{k\ge 1}\alpha_{k}. \]

\medskip

Let us consider the prediction problem  for the model (\ref{cov_equ}) based on observations $(Y_{1}, \mathcal{X}_{1}), \cdots, (Y_{n}, \mathcal{X}_{n})$ which is a trajectory of a stationary and ergodic process $(Y_{t}, \mathcal{X}_{t})_{t\in\mathbb{Z}}$, satisfying (\ref{cov_equ}) and (\ref{aut_equ}). 
The goal is to predict $Y_{n+1}$ from the observations $(Y_{n}, \mathcal{X}_{n}), \cdots, (Y_{n-p+1}, \mathcal{X}_{n-p+1}),$ for $p\in\mathbb{N}$. 
The learning theory with DNNs predictors developed above is performed with $X_{t}=\left((Y_{t-1}, \mathcal{X}_{t-1}), \cdots, (Y_{t-p}, \mathcal{X}_{t-p})\right), \mathcal{Y}\subset\mathbb{R}, $ $\mathcal{X}\subset (\mathbb{R}\times\mathbb{R}^{d_{x}})^{p}$ and a loss function $\ell:\mathbb{R}\times\mathcal{Y}\to[0, \infty)$.  
We focus on the class of DNNs estimators $\mathcal{H}_{\sigma}(L, N, B, F,S)$, for some $L,N,B,F,S \geq 0$.
The predictor obtained form the ERM algorithm is defined by,
\begin{equation}
  \widehat{h}_{n}=\underset{h\in\mathcal{H}_{\sigma}(L, N, B, F,S)}{\argmin}\widehat{R}_{n}(h), \; \text{with} \; \widehat{R}_{n}(h)= \frac{1}{n-p}\sum_{i=p+1}^{n}\ell(h(Y_{i-1}, \cdots, Y_{i-p}), Y_{i}).
\end{equation}

\medskip

 We assume that the activation function $\sigma$ is Lipschitz continuous; for instance, ReLU or sigmoid. Let us check the other assumptions of Theorem \ref{thm1} and Theorem \ref{thm2} for the class  models $\mathcal{AC-}X(\mathcal{M}, f)$.

\begin{enumerate}
\item[(i)] Under (\ref{Lipf_equ}), if the vector $(\eta_{0}, \xi_0)$ is bounded $a.s.$, then, one can show that (see for instance \cite{diop2022statistical}) $Y_0$ is $a.s.$ bounded and $\mathcal{Y}$ can be chosen bounded. Thus, \textbf{(A2)} holds for instance, for square loss. 
\item[(ii)]  Under the assumptions $\textbf{A}(f)$, $\textbf{A}(\mathcal{M})$ or $\textbf{A}(\mathcal{H})$, and (\ref{Lipf_equ}),
the process $(Y_{t}, \mathcal{X}_{t})_{t\in\mathbb{Z}}$ is $\theta$-weakly dependent with the coefficients $\theta(j)$ bounded as in (\ref{teta_bound_equ}).
Similar conditions to those of Subsection \ref{sub_bin_pred} can be obtained in:
\medskip
 
$\bullet$ the geometric case with,
\[ \alpha_{k, Y}(f) + \alpha_{k, Y}(\mathcal{M})+ \alpha_{k, Y}(H) + \alpha_{k}(g)= \mathcal{O}(a^{k}) \; \text{for some} \; a\in[0, 1) ,\]
and in:
\medskip 

$\bullet$ the Riemanian case with,
\[ \alpha_{k, Y}(f) + \alpha_{k, Y}(\mathcal{M})+\alpha_{k, Y}(H)+\alpha_{k}(g)=\mathcal{O}(k^{-\gamma}) \; \text{for some } \; \gamma > 1. \]
\end{enumerate}
Thus, if the activation and the loss functions are Lipschitz continuous and the innovations $(\eta_{t}, \xi_t)_{t\in\mathbb{Z}}$ are $a.s.$ bounded, the EMR algorithm for the prediction problem in  ARMAX, TARX, GARCH-X, APARCH-X, ARMAX-GARCH, type models, is consistent over the class of DNNs.

\section{Numerical results}
 In this section, we consider the prediction of binary time series by DNNs.
 
\subsection{Simulation study}
We consider a binary INGARCHX-X process $(Y_t, \mathcal{X}_t)_{t \in \Z}$ with values in $\{-1, 1 \} \times \R$ and satisfying 
\begin{equation}\label{Mod_sim}
 Y_{t}| \mathcal{F}_{t-1}\sim 2\mathcal{B}(p_{t})-1 \;\text{with} \; 2p_{t}-1=E[Y_{t}|\mathcal{F}_{t-1}]=f(Y_{t-1}, Y_{t-2}, \cdots; \mathcal{X}_{t-1}, \mathcal{X}_{t-2}, \cdots),
\end{equation}
with $\mathcal{F}_{t-1}=\sigma\left\{Y_{t-1}, \cdots; \mathcal{X}_{t-1}, \cdots\right\}$ and $f$ is with values in $[-1; 1]$.
Based on observations $(Y_{1}, \mathcal{X}_{1}), \cdots, (Y_{n}, \mathcal{X}_{n})$, our aims is to predict $Y_{n+1}$ from the DNN estimator as described in Subsections \ref{sub_bin_pred} and \ref{sub_bin_class}. This procedure is considered with $p=1$, $X_t =Y_{t-1}$ in the DGP1, and $p=2$, $X_t =(Y_{t-1}, Y_{t-2}, \mathcal{X}_{t-1})$ in the DGP2 below. 
We consider the following cases in (\ref{Mod_sim}):
\[
\begin{array}{ll}
\text{DGP1}: &  f(Y_{t-1},\ldots; \mathcal{X}_{t-1},\ldots) = -0.25 + 0.6 Y_{t-1} \\
\text{DGP2}: & f(Y_{t-1},\ldots; \mathcal{X}_{t-1},\ldots) = 0.1 - 0.15 \max(Y_{t-1},0) + 0.25 \min(Y_{t-1},0) + 0.15 Y_{t-2} + 0.2\dfrac{1}{ 1 +  \mathcal{X}_{t-1}^2},
\end{array} 
\] 
where $(\mathcal{X}_t)_{t \in \Z}$ is an AR(1) process.
The DGP1 is related to the real data example in the subsection below. 
In the sequel, $h_0$ denote the predictor obtained from the true DGP; that is, 
\begin{equation} \label{def_h0}
 h_0(X_t) = \text{sign}(f(X_t))  \text{ for all } t \in \Z.
\end{equation}
 Note that, $h_0$ is the Bayes classifier according to the $0-1$ loss function; that is 
 $R_{01}(h_0) = \inf_{h \in \mathcal{F}(\mathcal{X}, \mathcal{Y})} R_{01}(h) $,
 where $R_{0,1}$ is defined at (\ref{risk_01}) and $\mathcal{F}(\mathcal{X}, \mathcal{Y})$ is the set of measurable functions from $\mathcal{X}$ to $\mathcal{Y}$.

\medskip

 For both DGPs, we used a network architecture of 2 hidden layers with 16 hidden nodes for each layer and the ReLU activation function ($\sigma(x) = \max(x,0)$). The network weights were trained in the R package Keras, by using the algorithm Adam (\cite{kingma2014adam}) with learning rate $10^{-3}$ and the minibatch size of 32.
 We stopped training when the accuracy is not improved in 30 epochs.
 The hinge loss is used ($\phi(z)=\max(1-z,0)$) and the $tanh$ activation function is used for the output layer. We predict -1 if the output is negative and predict 1 otherwise.   
Denote by $\mathcal{H}_\sigma$ the set of the network architecture considered and by $\Theta$ its parameter space.

\medskip
 
\begin{figure}[h!]
\begin{center}
\includegraphics[height=12.5cm, width=18.5cm]{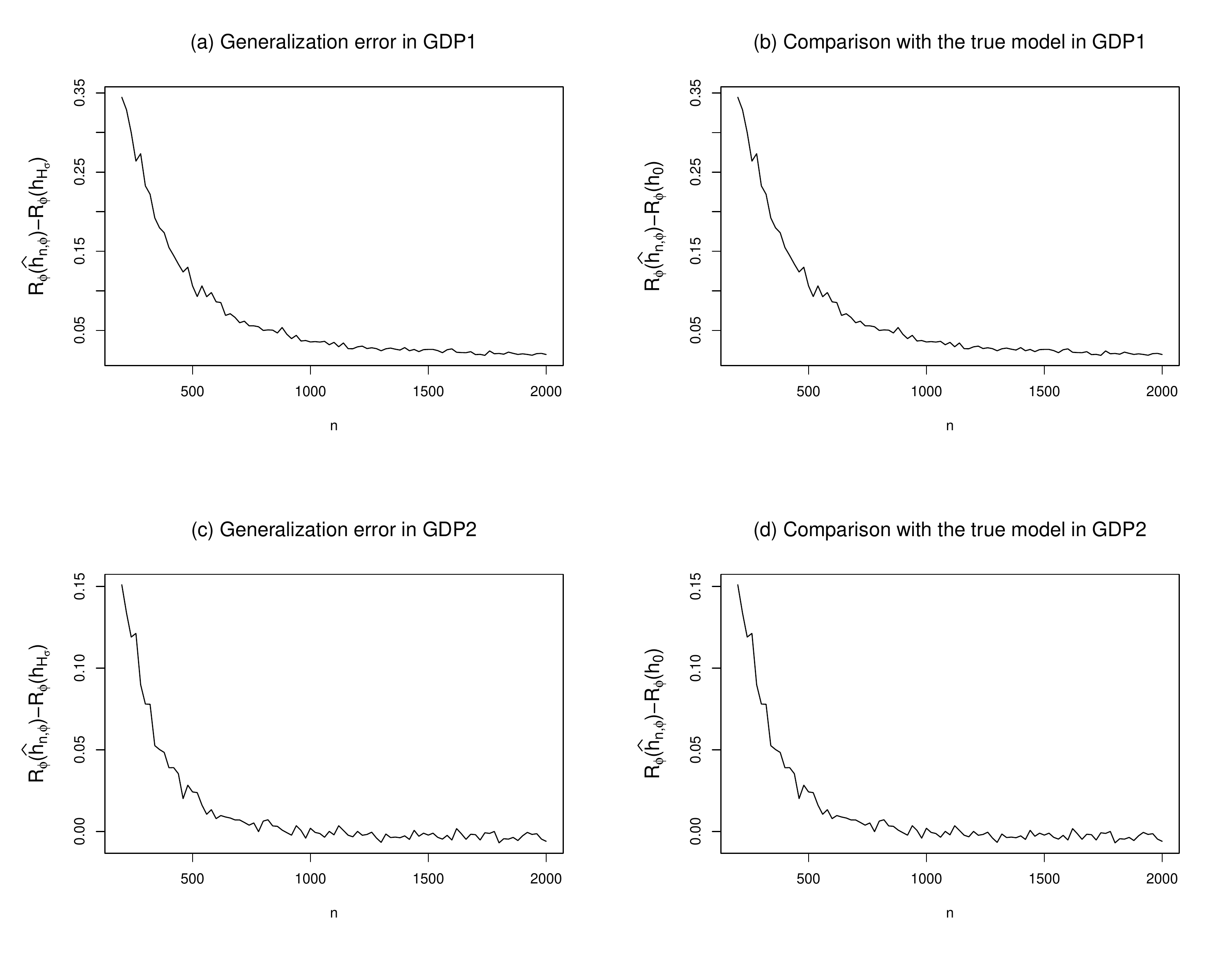}
\end{center}
\vspace{-.7cm}
\caption{\it Plots of $\big(n,R_\phi(\widehat{h}_{n,\phi}) - R_\phi(h_{\mathcal{H}_\sigma}) \big)$ and $\big(n,R_\phi(\widehat{h}_{n,\phi}) - R_\phi(h_{0}) \big)$ with $n$ from 200 to 2000 in DGP1 (a), (b) and DGP2 (c), (d).}
\label{Graphe_Sim_DGP_1_2}
\end{figure}

Firstly, for each DGP considered, a sample $(Y_0', \mathcal{X}_0'), (Y_1', \mathcal{X}_1'), (Y_2', \mathcal{X}_2'),\ldots, (Y_m', \mathcal{X}_m')$ with $m=10^4$ is generated and the target network $h_{\mathcal{H}_\sigma}$ (see (\ref{def_target_DNN})) is estimated by $\widetilde{h}$ where,
\[ \theta(\widetilde{h}) = \underset{\theta(h) \in \Theta}{\argmin}\widetilde{R}_{\phi,1}(h), \text{ with } \widetilde{R}_{\phi,1}(h) := \dfrac{1}{m} \sum_{i=1}^m \max\big(1- Y_i' h(X_i'), 0 \big) \text{ and } X_i'=(Y_{i-1}', \mathcal{X}_{i-1}').\]
Therefore, an estimation of $R_\phi(h_{\mathcal{H}_\sigma})$ is $\widetilde{R}_{\phi,1}( \widetilde{h} )$. 
Also, the risk of $h_0$ (see (\ref{def_h0})) is estimated by  $\widetilde{R}_{\phi,1}( h_0 )$.
Secondly, for $n=200,220,240,\ldots,2000$, a trajectory $((Y_1, \bchi_1), (Y_2, \mathcal{X}_2), \ldots, (Y_n, \mathcal{X}_n))$ is generated from the true DGP. $\widehat{h}_{n,\phi}$ is estimated with
\[ \theta(\widehat{h}_{n,\phi}) = \underset{\theta(h) \in \Theta}{\argmin}\widehat{R}_{n,\phi}(h) .\] 
Thus, an estimation of $R(\widehat{h}_{n,\phi})$ is $\widetilde{R}_2 (  \widehat{h}_{n,\phi}  )$, with, 
\[ \widetilde{R}_{\phi,2}(h) = \dfrac{1}{n} \sum_{i=1}^n \max\big( 1-Y_i'' h(X_i''), 0\big) \text{ for } \theta(h) \in \Theta \text{ and } X_i''=(Y_{i-1}'', \mathcal{X}_{t-1}''), \] 
where $((Y_0'', \mathcal{X}_0''), (Y_1'', \mathcal{X}_1''), (Y_2'', \mathcal{X}_2''),\ldots, (Y_n'', \mathcal{X}_n''))$ is generated from the true DGP and is independent of the sample used for the estimation of $\widehat{h}_{n,\phi}$. 
For each fixed $n$, the Monte Carlo estimation of $R(\widehat{h}_{n,\phi})$ is based on 500 replications.  
 Figure \ref{Graphe_Sim_DGP_1_2} displays the curves of the points $\big(n,R_\phi(\widehat{h}_{n,\phi}) - R_\phi(h_{\mathcal{H}_\sigma}) \big)$ and $\big(n,R_\phi(\widehat{h}_{n,\phi}) - R_\phi(h_{0}) \big)$ in DGP1 and DGP2.
For both DGP1 and DGP2, one can see that, the values of $R_\phi(\widehat{h}_{n,\phi}) - R_\phi(h_{\mathcal{H}_\sigma})$ approaching zero as the sample size $n$ increases. 
 These numerical findings are in accordance with the consistency of
the ERM algorithm within the class of DNNs predictors and the generalization bound in Theorem \ref{thm1} and \ref{thm2}. 
 One can also see that, as $n$ increases, the risk of the neural network predictor $\widehat{h}_{n,\phi}$ approaches that of the predictor $h_0$ obtained from the true model.

\subsection{Application to the US recession data}
\begin{figure}[h!]
\begin{center}
\includegraphics[height=8.5cm, width=10.5cm]{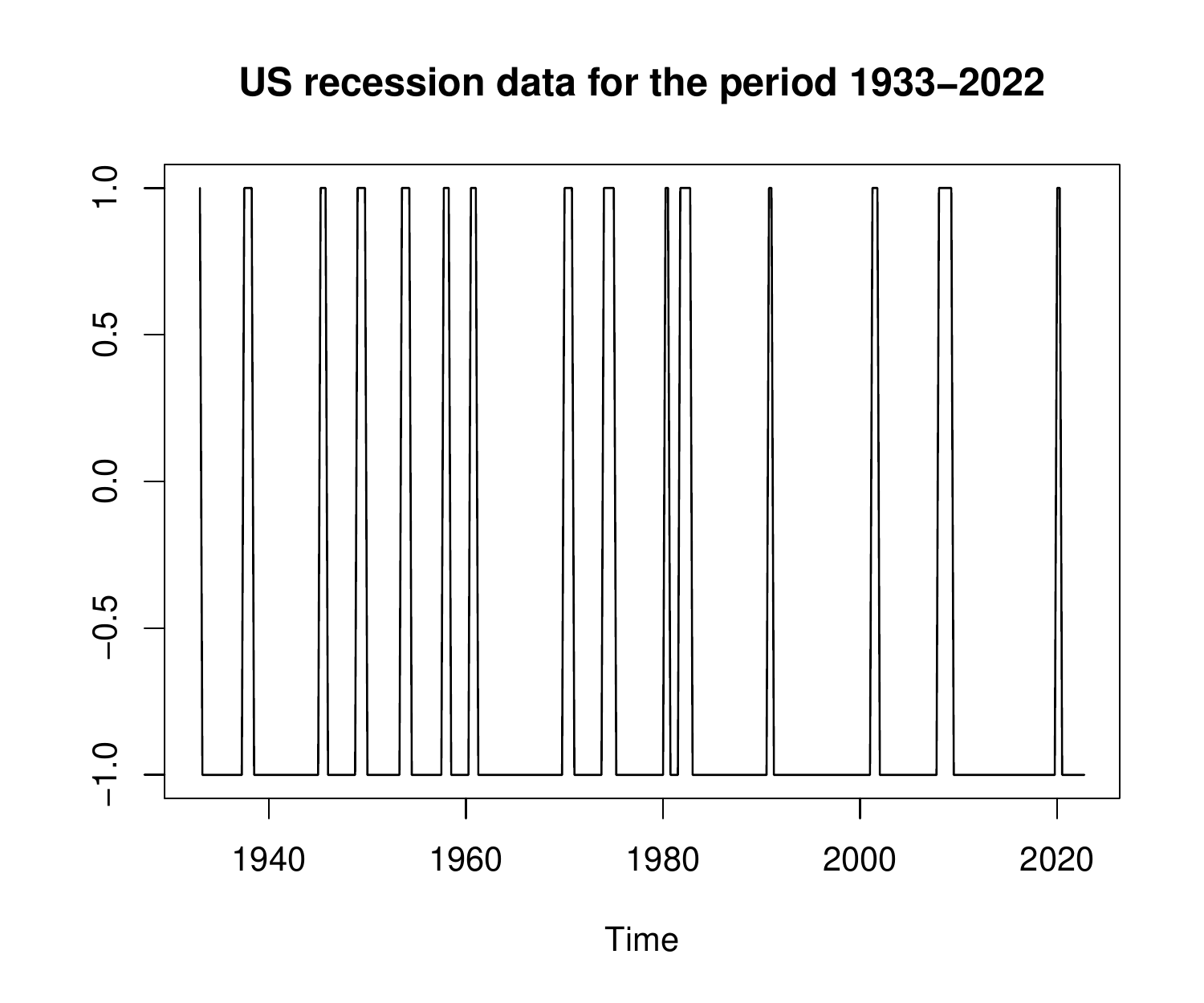}
\end{center}
\vspace{-.7cm}
\caption{\it Quarterly recession data from the USA for the period 1933-2022.}
\label{Graph_US_rec}
\end{figure}

  We consider the series of the quarterly recession data from the USA for the period 1933-2022, there are 360 observations.
  These data are available at https://fred.stlouisfed.org/series/USRECQ , and represent a binary variable that is equal to 1 if there is a recession in at least one month in the quarter and 0 otherwise. 
 We have recoded the original data into $1$ if there is a recession in at least one month in the quarter and $-1$ otherwise, see Figure \ref{Graph_US_rec}.

  These data have already been analyzed by several authors, see for instance \cite{hudecova2013structural}, \cite{diop2017testing}, \cite{diop2021piecewise}. 
  These last two works have found that, one order autoregression allows to well fit these data. That is, the model (\ref{Mod_equ1}), where 
  \[ f(Y_{t-1}) = \alpha_0 + \alpha_1 Y_{t-1}, \]
 %
with $|\alpha_0| + |\alpha_1| < 1$.
The maximum likelihood estimation (see for instance \cite{davis2016theory}, \cite{diop2020density}) of the parameter $\theta=(\alpha_0, \alpha_1)$ is $\widehat{\theta}=(-0.248,0.660)$.
This real data example is close to the DGP1 in the subsection above.
For prediction problem, we consider the network architecture of the previous subsection (2 hidden layers with 16 hidden nodes for each layers and the ReLU activation function) and with input variable $X_t=Y_{t-1}$.
The network is trained on the first half of the data and the second half (used as test data) is used to evaluate its risk and accuracy.
Table \ref{CM_rec} provides the confusion matrix obtained from the test data.
\begin{table}[h!] 
\centering 
\begin{tabular}{@{}llwr{2.2em}wc{2em}}
\toprule
                  & & \multicolumn{2}{l}{\thead{\makebox[0pt]{Predicted}\\ value}}\\ \midrule
                  & & -1 & 1 \\
\multirow{2}{*}{Actual value } & -1 & 153 & 6 \\
                                            & 1 & 5 & 15\\ \bottomrule
\end{tabular}
\caption{\small \it Confusion matrix obtained from the test data.}
\label{CM_rec}

\end{table}
The risk (with respect to the hinge loss ) and the accuracy are 0.2445 and 0.9385 respectively (note that, the risks of the target neural network $h_{\mathcal{H}_\sigma}$ and that of the predictor $h_0$ at (\ref{def_h0}) in DGP1 are 0.2288979 and 0.2288229 respectively).
One can also see that, the accuracy of predicting the recession period is 0.75.
These results show that, the neural networks work overall well for these data. So, if the network is well trained, with only the data of this quarter, one can predict with a quite satisfactory accuracy whether or not the next quarterly will be a recession period.

\section{Proofs of the main results}
\subsection{Proof of Proposition \ref{prop}}
Consider a class of DNNs predictors $\mathcal{H}_{\sigma}(L, N, B, F,S)$ with $L, N, B, F, S \geq 0$.
It suffices to show that, the conditions of Theorem 3.2 and 3.4 in \cite{diop2022statistical} are satisfied for the class $h\in\mathcal{H}_{\sigma}(L, N, B, F,S)$, the loss $\ell:\mathbb{R}\times\mathcal{Y}\to [0, \infty)$ and the process $\{Z_{t}=(X_{t},Y_{t}), t\in\mathbb{Z} \}$.
By the definition of the set  $ \mathcal{H}_{\sigma}(L, N, B, F,S)$, it holds that, $\underset{h\in\mathcal{H}_{\sigma}(L, N, B, F,S)}{\sup} \|h \|_{\infty} \leq F < \infty $.
Therefore, according to the assumptions \textbf{(A2)} and \textbf{(A3)}, it remains to show that, there exists $\mathcal{K}_{\mathcal{H}_\sigma} >$ such that, for all $h \in \mathcal{H}_{\sigma}(L, N, B, F,S)$, $h$ is $\mathcal{K}_{\mathcal{H}_\sigma}$-Lipschitz. 

\medskip 

 Let $h\in\mathcal{H}_{\sigma}(L, N, B, F,S)$, for all $x, x' \in \R^d$,  we have from \textbf{(A1)},
\begin{align}
\nonumber \|h(x)-h(x')\| & = \|A_{L+1}\circ\sigma_{L}\circ A_{L}\circ\sigma_{L-1}\circ \cdots \circ\sigma_{1}\circ A_{1}(x)-A_{L+1}\circ\sigma_{L}\circ A_{L}\circ\sigma_{L-1}\circ \cdots \circ\sigma_{1}\circ A_{1}(x') \| \\
 \nonumber & \leq C_{\sigma}^{L}\prod_{i=1}^{L+1} \; \| W_{i}\| \; \|x-x' \| \\
 \nonumber & \leq C_{\sigma}^{L} B^{L+1} \|x-x' \|,
\end{align}
where $\|\cdot\|$ denotes any vector, matrix norm.
With $\mathcal{K}_{\mathcal{H}_\sigma} = C_{\sigma}^{L} B^{L+1}$, the conditions of Theorem 3.2 and 3.4 in \cite{diop2022statistical} holds and Proposition \ref{prop} follows.

\subsection{Proof of Theorem \ref{thm1}}\label{P_prop1}
(i) Let $\varepsilon>0$ and set $\delta=\frac{\varepsilon}{4G}>0$.
  Since  the activation function $\sigma$ is $C_{\sigma}$-Lipschitz, we have from Proposition 1 in \cite{ohn2019smooth},
\begin{equation}\label{P_inqu1}
 \log\mathcal{N}(\mathcal{H}_{\sigma}(L, N, B, F,S),\frac{\varepsilon}{4G})\leq 2L(S+1)\log\left(\frac{4G}{\varepsilon}C_{\sigma}L(N+1)(B\lor 1)\right).
\end{equation}
It is equivalent to  \\
\begin{equation}\label{P_inqu2}
\mathcal{N}(\mathcal{H}_{\sigma}(L, N, B, F,S),\frac{\varepsilon}{4G})\leq\exp\left(2L(S+1)\log\left(\frac{4G}{\varepsilon} C_{\sigma}L(N+1)(B\lor 1)\right)\right).
\end{equation}
Note that, if $\varepsilon > 2M$, then $ P\left\{\sup_{h\in\mathcal{H}_{\sigma}(L,N,B,F)}[R(h)-\widehat{R}_{n}(h)] > \varepsilon\right\}=0.$  For $\varepsilon\in (0,2M]$, from Proposition 3.1 and Remark 3.3  in  $\cite{diop2022statistical}$, by taking $ A_{n}=2nM^{2}\Psi (1, 1)L_{1}$ and $ B_{n}= 2ML_{2}\max (2^{3+\mu}/ \Psi (1, 1), 1)$  we have,
\begin{equation}\label{Mod_1}
  P\left\{\sup_{h\in\mathcal{H}_{\sigma}(L, N, B, F,S)}[R(h)-\widehat{R}_{n}(h)]>\varepsilon\right\}\leq\exp\left(- \frac{n^{2}\varepsilon^{2}/4}{C_{1}n + 2C_{2}^{1/(\mu+2)}(n\varepsilon/2)^{(2\mu+3)/(\mu+2)}}\right).
\end{equation}
In addition to Proposition \ref{prop} and (\ref{P_inqu2}), it holds that,
\begin{multline}\label{Mod_2}
 P\left\{\sup_{h\in\mathcal{H}_{\sigma}(L, N, B, F,S)}[R(h)-\widehat{R}_{n}(h)] > \varepsilon\right\} \\
  \leq\exp\left(2L(S+1)\log\left(\frac{4G}{\varepsilon}C_{\sigma}L(N+1)(B\lor 1)\right) - \frac{n^{2}\varepsilon^{2}/4}{C_{1}n + 
 2C_{2}^{1/(\mu+2)}(nM)^{(2\mu+3)/(\mu+2)}}\right), 
\end{multline}
with  $ C_{1} = 4M^{2}\Psi(1, 1)L_{1} \; \text{and} \;  C_{2}= 2ML_{2}\max(\frac{2^{3 + \mu}}{\Psi(1, 1)}, 1)$. 
Let $ 0 < \eta < 1$. Consider the following equation with respect to $\varepsilon$:
\[ \exp\left(2L(S+1)\log\left(\frac{4G}{\varepsilon}C_{\sigma}L(N+1)(B\lor 1)\right) - \frac{n^{2}\varepsilon^{2}/4}{C_{1}n + 2C_{2}^{1/(\mu+2)}(nM)^{(2\mu+3)/(\mu+2)}}\right) =\eta .\] 
It is equivalent to,
\begin{equation}\label{Mod_3}
 2L(S+1)\log\left(\frac{4G}{\varepsilon}C_{\sigma}L(N+1)(B\lor 1)\right) - \frac{n^{2}\varepsilon^{2}/4}{C_{1}n + 2C_{2}^{1/(\mu+2)}(nM)^{(2\mu+3)/(\mu+2)}} +\log(1/ \eta) =0 .
\end{equation}
Set,
\[ C_{n, 1}=\frac{n^{2}}{4C_{1}n + 8C_{2}^{1/(\mu +2)}(nM)^{(2\mu +3)/(\mu + 2)}}. \]
We get,
\[2L(S+1)\log\left(\frac{4G}{\varepsilon}C_{\sigma}L(N+1)(B\lor 1)\right) - C_{n, 1}\varepsilon^{2} + \log(1/\eta)=0. \]
i.e. 
\[ -2L(S+1)\log\varepsilon  + C_{6} - C_{n, 1}\varepsilon^{2} - \log(\eta)=0,\]
with $C_{6} =2L(S+1)\log\left(4 G C_{\sigma}L(N+1)(B\lor 1)\right)$. 
i.e.
\[  2L(S+1)\log\varepsilon + C_{n, 1}\varepsilon^{2} +\log(\eta) - C_{6}=0. \]
Consider the function $\phi(\varepsilon)=2L(S+1)\log\varepsilon + C_{n, 1}\varepsilon^{2} +\log(\eta) - C_{6},$ for $\varepsilon\in(0, 2M)$.
We have, $\phi(\varepsilon)\to\ -\infty $ as $\varepsilon\to 0$.
\begin{align}\label{C_n_equ1}
\phi(2M)>0 \nonumber &\Rightarrow 2L(S+1)\log(2M) + 4C_{n, 1}M^{2} +\log(\eta) - C_{6}>0\\
        & \Rightarrow C_{n, 1}>\frac{1}{4M^{2}}(C_{6} - 2L(S+1)\log(2M) -\log(\eta)).
\end{align}

Since $(2\mu +3)/(\mu+2)>1,$ one can easily see that
\begin{equation}\label{C_n_1}
 C_{n, 1}\ge \frac{n^{2}}{n^{(2\mu +3)/(\mu+2)}(4C_{1} + 8C_{2}^{1/(\mu +2)}M^{(2\mu +3)/(\mu + 2)})}=\frac{n^{\frac{1}{\mu+2}}}{C_{4}},
\end{equation}
with $ C_{4}=4C_{1} + 8C_{2}^{1/(\mu +2)}M^{(2\mu +3)/(\mu + 2)}$.
Thus, to get (\ref{C_n_equ1}), it suffices that\\
\[ \frac{n^{\frac{1}{\mu+2}}}{C_{4}}>\frac{1}{4M^{2}}(C_{6} - 2L(S+1)\log(2M) -\log(\eta))_{+}\ge\frac{1}{4M^{2}}(C_{6} - 2L(S+1)\log(2M) -\log(\eta)) .\]
Recall that, $(x)_{+}=\max(x, 0)$ for all $ x \in \R$. 
That is, 
\begin{equation}\label{n_1}
    n>\left(\frac{C_{4}}{4M^{2}}(C_{6} - 2L(S+1)\log(2M) -\log(\eta))_{+}\right)^{\mu+2}.
\end{equation}
The function $\varepsilon\mapsto\phi(\varepsilon)$ is strictly increasing and under the condition (\ref{n_1}), we get $\phi(\varepsilon)\to -\infty$ as $\varepsilon\to 0$ and $\phi(2M)>0.$ Thus, there exists a unique $\varepsilon_1(n, \eta)\in(0, 2M),$  such that $\phi(\varepsilon(n, \eta))=0$.
Set $\varepsilon_{0}=\frac{2M}{n^{\frac{1}{\alpha(\mu+2)}}}$ for $\alpha  >2$.
We have in addition to (\ref{C_n_1}),
\begin{align*}
\phi(\varepsilon_{0})  \nonumber &=2L(S+1)\log\varepsilon_{0} + C_{n, 1}\varepsilon_{0}^{2} +\log(\eta) - C_{6}\\
  & \geq  2L(S+1)\log(2M) - \frac{2L(S+1)}{\alpha(\mu+2)}\log n + \frac{4{M}^{2}}{C_{4}}n^{\frac{1}{\mu+2} (1-\frac{1}{\alpha})} + \log(\eta) - C_{6} \\
 & \ge 2L(S+1)\log(2M) - \frac{2L(S+1)}{\alpha(\mu+2)}\log n + \frac{4{M}^{2}}{C_{4}}n^{\frac{\alpha-2}{\alpha(\mu+2)}} + \log(\eta) - C_{6}.
\end{align*}
To get $\phi(\varepsilon_{0})>0,$  it suffices that  $2L(S+1)\log(2M) - \frac{2L(S+1)}{\alpha(\mu+2)}\log n + \frac{4{M}^{2}}{C_{4}}n^{ \frac{\alpha-2}{\alpha(\mu+2)} } + \log(\eta) - C_{6} > 0$, \\
i.e.
\begin{equation}\label{n'_mu_1}
 n^{ \frac{\alpha-2}{\alpha(\mu+2)} }\frac{4{M}^{2}}{C_{4}}\left(1 - \frac{2C_{4}L(S+1)}{\alpha 4{M}^{2}(\mu+2)}\frac{\log n}{n^{  \frac{\alpha-2}{\alpha(\mu+2)} }}\right)> C_{6} - \log(\eta) - 2L(S+1)\log(2M)
\end{equation}
Since,
\[   \frac{2C_{4}L(S+1)}{\alpha 4{M}^{2}(\mu+2)}\frac{\log n}{n^{  \frac{\alpha-2}{\alpha(\mu+2)} }} \limiten 0,  \]
 there exists $n_0(L, S,\alpha, M, \mu, C_4) >0$ such that, 
 \begin{equation} \label{def_n_0}
  n > n_0(L, S,\alpha, M, \mu, C_4) \Rightarrow  \frac{2C_{4}L(S+1)}{\alpha 4{M}^{2}(\mu+2)}\frac{\log n}{n^{  \frac{\alpha-2}{\alpha(\mu+2)} }} < \frac{1}{2} .
\end{equation} 
Thus, to get (\ref{n'_mu_1}), with $n >n_0(L, S,\alpha, M, \mu, C_4) $ it suffices that,
\[ n^{ \frac{\alpha-2}{\alpha(\mu+2)}}\frac{4{M}^{2}}{2C_{4}}> C_{6} - \log(\eta) - 2L(S+1)\log(4M). \]
That is, 
\[  n > \left( \frac{C_{4}}{2{M}^{2}} (C_{6} - \log(\eta) - 2L(S+1)\log (2M))_{+} \right)^{ \frac{ \alpha (\mu+2)}{\alpha-2} }. \]
Thus, for sufficiently large $n$, the unique solution $\varepsilon_{1}(n, \eta, \alpha)$ of $\phi(\varepsilon(n, \eta))=0$,  satisfies $\varepsilon_{1}(n, \eta, \alpha)<\frac{2M}{n^{\frac{1}{\alpha(\mu+2)}}}< 2M$ for $\alpha>2$.
Hence,
from (\ref{Mod_1}), it holds that, with probability at least $1 - \eta,$
\[ \sup_{h\in\mathcal{H}_{\sigma}(L, N, B, F,S)}[R(h)-\widehat{R}_{n}(h)]\leq\varepsilon_{1}(n, \eta, \alpha); \]
which implies,
\begin{equation}\label{Reps_eq}
    R(\widehat{h}_{n})-\widehat{R}_{n}(\widehat{h}_{n})\leq\varepsilon_{1}(n, \eta, \alpha).
\end{equation}
This establishes the first part of the Theorem \ref{thm1}.

\medskip
 
(ii) From the Proposition 3.1 in \cite{diop2022statistical} , we have the following inequality
 for all $h\in\mathcal{H}_{\sigma}(L, N, B, F,S)$,
 \begin{equation}\label{Mod_equ2}
 P\left\{\widehat{R}_{n}(h) - R(h)>\varepsilon\right\}\leq\exp\left(-\frac{n^{2}\varepsilon^{2}/2}{A_{n} + B_{n}^{1/(\mu+2)}(n\varepsilon)^{(2\mu+3)/(\mu+2)}}\right).
\end{equation}
Thus, by taking $ A_{n}=2nM^{2}\Psi (1, 1)L_{1}$ and $ B_{n}= 2ML_{2}\max (2^{3+\mu}/ \Psi (1, 1), 1)$ as above,  with the target neural network $ h_{\mathcal{H}_{\sigma}(L, N, B, F,S)}$, it comes that,
\begin{align}\label{Mod_equ3}
P\left\{\widehat{R}_{n}(h_{\mathcal{H}_{\sigma}(L, N, B, F,S)}) - R(h_{\mathcal{H}_{\sigma}(L, N, B, F,S)})>\varepsilon\right\}  \nonumber & \leq\exp\left(-\frac{n^{2}\varepsilon^{2}/2}{A_{n} + B_{n}^{1/(\mu+2)}(n\varepsilon)^{(2\mu+3)/(\mu+2)}}\right)\\
  & \leq\exp\left(-\frac{n^{2}\varepsilon^{2}}{C_{1}n + 2C_{2}^{1/\mu+2}(2nM)^{(2\mu+3)/(\mu+2)}}\right)\\
 \nonumber & \leq\exp(-C'_{n}\varepsilon^{2}),
 \end{align}
 where $C'_{n}=\frac{n^{2}}{C_{1}n + 2C_{2}^{1/\mu+2}(2nM)^{(2\mu+3)/(\mu+2)}}$. 
 Consider the equation, with respect to $\varepsilon$, 
 \[ \exp\left(-C'_{n}\varepsilon^{2}\right)= \eta .\]
 A solution of this equation is,
 \[ \varepsilon_{1}'(n, \eta)=\left[\frac{\log(1/ \eta)}{C'_{n}}\right]^{\frac{1}{2}} .\]
 Thus,  from  (\ref{Mod_equ2}),  we have,  
 \[ \widehat{R}_{n}(h_{\mathcal{H}_{\sigma}(L, N, B, F,S)}) - R(h_{\mathcal{H}_{\sigma}(L, N, B, F,S)})\leq\varepsilon_{1}'({n, \eta}) \]
 with a probability at least $ 1-\eta.$ 
 Since $\widehat{R}_{n}(\widehat{h}_{n})\leq\widehat{R}_{n}(h_{\mathcal{H}_{\sigma}(L, N, B, F,S)})$ (from the definition of $\widehat{h}_{n}=\underset{h\in\mathcal{H}_{\sigma}(L, N, B, F,S)}{\argmin}[\widehat{R}_{n}(h)]$), we deduce  
 \begin{equation}\label{Mod_equ4}
\widehat{R}_{n}(\widehat{h}_{n})-R(h_{\mathcal{H}_{\sigma}(L, N, B, F,S)})\leq\varepsilon_{1}'({n, \eta}).
 \end{equation}
Therefore, according to (\ref{Reps_eq}) and (\ref{Mod_equ4}), it holds, with probability at least $1-2\eta$, that, 
\[ R(\widehat{h}_{n}) - R(h_{\mathcal{H}_{\sigma}(L, N, B, F,S)})\leq\varepsilon_{1}(n, \eta,\alpha)+\varepsilon_{1}'(n, \eta). \]
This completes the proof of the Theorem.

\subsection{Proof of Theorem \ref{thm2}}
(i) According to (\ref{E_equ6}), take $A'_{n}=nC$,  with $B'_{n}=\log\left(\frac{n}{n^{1/4}C}\right)$. 
 Thus, for $\varepsilon \in (0, 2M ]$, by using (\ref{P_equ5}) and (\ref{P_inqu2}), we get for $n$ large enough,
\begin{multline}\label{P_equ33}
P\left\{\underset{h\in\mathrm{H_{\sigma}}(L, N, B, F,S)}{\sup}[R(h) - \widehat{R}_{n}(h)]>\varepsilon\right\}  \leq C_{3}\exp\left(2L(S+1)\log\left(\frac{4G}{\varepsilon} C_{\sigma}L(N+1)(B\lor 1)\right)\right) \\
 \times \exp\left(\log\log n - \frac{n^{2}\varepsilon^{2}/4}{nC +  \log n \; n^{\nu - 1/4}M^{\nu}/C }\right).
\end{multline}
Let us consider the following equation with respect to $\varepsilon$:
\[ C_{3}\exp\left(2L(S+1)\log\left(\frac{4G}{\varepsilon} C_{\sigma}L(N+1)(B\lor 1)\right)+\log\log n - \frac{n^{2}\varepsilon^{2}/4}{nC +  \log n \; n^{\nu - 1/4}M^{\nu}/C }\right)=\eta, \] 
i.e.
\[ 2L(S+1)\log\left(\frac{4G}{\varepsilon} C_{\sigma}L(N+1)(B\lor 1)\right)+\log\log n - \frac{n^{2}\varepsilon^{2}/4}{nC +  \log n \; n^{\nu - 1/4}M^{\nu}/C } - \log(\eta) + \log C_{3} = 0, \] 
i.e.
\[ -2L(S+1)\log \varepsilon + 2L(S+1)\log\left(4G C_{\sigma}L(N+1)(B\lor 1)\right) +  \log\log n - \frac{n^{2}\varepsilon^{2}/4}{nC +  \log n \; n^{\nu - 1/4}M^{\nu}/C } - \log(\eta) + \log C_{3} = 0, \] 
i.e.
\begin{equation}\label{proof_eq_egal_0}
-2L(S+1)\log \varepsilon + 2L(S+1)\log\left(4G C_{\sigma}L(N+1)(B\lor 1)\right) + \log\left(C_{3}\log n/ \eta \right)- \frac{n^{2}\varepsilon^{2}/4}{nC +  \log n \; n^{\nu - 1/4}M^{\nu}/C } = 0 . 
\end{equation}
Set,
 \[ C_{n, 2}=\frac{n^{2}/4}{nC +  \log n \; n^{\nu - 1/4}M^{\nu}/C }. \]
We get from (\ref{proof_eq_egal_0}),
\[ 2L(S+1)\log \varepsilon + C_{n, 2}\varepsilon^{2}  + \log(\eta/C_{3})  - C_{6} -\log(n) = 0, \]
with $C_{6} =2L(S+1)\log\left(4G C_{\sigma}L(N+1)(B\lor 1)\right)$.
Consider the function 
\[ \varphi(\varepsilon)=2L(S+1)\log \varepsilon + C_{n, 2}\varepsilon^{2}  + \log(\eta/C_{3})  - C_{6} -\log(n) \text{ for }  \varepsilon\in (0, 2M] . \]
 We have $\varphi(\varepsilon) \to -\infty$ as $\varepsilon\to 0$. 
 Also,
\begin{align}\label{Prop_equ3.6}
 \nonumber \varphi(2M) > 0 & \Rightarrow 2L(S+1)\log (2M) + 4 C_{n, 2}M^{2}  + \log(\eta/C_{3})  - C_{6} -\log(n) > 0 \\
 & \Rightarrow C_{n, 2} -\frac{\log(n)}{4M^{2}} > \frac{1}{4M^{2}}\left(C_{6} - 2L(S+1)\log (2M) - \log(\eta/C_{3})\right)
\end{align}
Note that, since $\nu\in (0, 1)$, for $n$ large enough, we have 
\begin{equation}\label{Prop1_3.6}
C_{n, 2} \ge \frac{n^{2}/4}{C'_{1}n + C'_{2}nM^{\nu}} \ge \frac{n}{C_{5}} \text{ and }  \frac{\log(n)}{4M^{2}} < \frac{n}{2 C_{5}}  
\end{equation}
with $ C_{5}= 4(C + M^{\nu}/C)$.
Thus, to get  (\ref{Prop_equ3.6}), it suffices that,
\[ \frac{n}{2 C_{5}} > \frac{1}{4M^{2}}\left(C_{6} - 2L(S+1)\log (2M) - \log(\eta/C_{3})\right)_{+} \ge \frac{1}{4M^{2}}\left(C_{6} - 2L(S+1)\log (2M) - \log(\eta/C_{3})\right) .\]
That is,
\begin{equation}\label{prop2_equ3.6}
n > \frac{C_{5}}{8M^{2}}\left(C_{6} - 2L(S+1)\log (2M) - \log(\eta/C_{3})\right)_{+} .
\end{equation}
Therefore, the function $\varepsilon \mapsto \varphi(\varepsilon)$ is strictly increasing and for sufficiently large $n$ and satisfying (\ref{prop2_equ3.6}), we get $\varphi(\varepsilon) \to -\infty$ as $\varepsilon \to 0$ and $ \varphi(2M) > 0$. Thus, there exists a unique  $\varepsilon(n, \eta) \in (0, 2M)$, such that $\varphi(\varepsilon(n, \eta))=0$.
Set $\varepsilon'_{0}=\dfrac{2M}{n^{\frac{1}{\alpha}}}$  for  $\alpha > 2$.
In the same way as in the proof of the first part of Theorem \ref{thm1}, one find that, 
to get  $\varphi(\varepsilon'_{0}) > 0$ it suffices that,
\begin{equation}\label{cond_n_large}
 n > \left(\frac{C_{5}}{2M^{2}}(C_{6} - \log(\eta / C_{3}) - 2L(S+1)\log(2M))_{+}\right)^{\alpha/(\alpha-2)} .
\end{equation}
Thus, for sufficiently large $n$ and satisfying (\ref{cond_n_large}),
$\varphi(\varepsilon_{0}') > 0$. 
Thus, the unique solution $\varepsilon_{2}(n, \eta, \nu, \alpha)$ of $\varphi(\varepsilon) = 0$ satisfies $\varepsilon_{2}(n, \eta, \nu, \alpha) < \frac{2M}{n^{\frac{1}{\alpha }}}$ for  $\alpha > 2$.\\
According to (\ref{P_equ33}), we deduce that with probability  at least $1-\eta$,
\begin{equation}\label{F1_eq}
    R(\widehat{h}_{n}) - \widehat{R}(\widehat{h}_{n})\leq \varepsilon_{2}(n, \eta , \nu, \alpha).
\end{equation}
Hence, the part (i) is established.

\medskip

(ii) In the same way,  we can go along similar lines as in (\ref{Mod_equ4}) (see proof of Theorem \ref{thm1}(ii)) to establish that,  with probability at least $1-\eta$
\begin{equation}\label{F2_eq}
    \widehat{R}_{n}(\widehat{h}_{n}) - R(h_{\mathcal{H}_{\sigma}(L,N,B,F)})\leq \varepsilon'_{2}(n, \eta, \nu),
\end{equation}
where,
\[ \varepsilon'_{2}(n, \eta, \nu)=\left[\frac{\log(C_{3}\log n/ \eta)}{C'_{n, 2}}\right]^{\frac{1}{2}} \text{ with } C'_{n, 2}= \frac{n^{2}}{nC + \log n \; n^{\nu -1/4}M^{\nu}/C} .\] 
Combining (\ref{F1_eq}) and (\ref{F2_eq}), with probability at least $ 1 - 2\eta$, we get 
$$R(\widehat{h}_{n}) -  R(h_{\mathcal{H}_{\sigma}(L, N, B, F,S)})\leq\varepsilon_{2}(n, \eta, \nu, \alpha) +\varepsilon'_{2}(n, \eta, \nu)$$
Which completes the proof  of the Theorem.

\end{document}